\documentclass[11pt]{article}
\usepackage[margin=1in]{geometry}
\usepackage{times}

\usepackage{amsmath,amssymb,amsfonts,amsthm,mathtools}
\usepackage[utf8]{inputenc}
\usepackage[T1]{fontenc}
\usepackage[numbers,sort&compress]{natbib}
\usepackage{hyperref}
\usepackage{url}
\usepackage{booktabs}
\usepackage{nicefrac}
\usepackage{microtype}
\usepackage{xcolor}
\DeclareRobustCommand{\revone}[1]{#1}
\usepackage{graphicx}
\usepackage{float}
\usepackage{subcaption}
\usepackage{multirow}
\usepackage{enumitem}

\usepackage[capitalize,noabbrev]{cleveref}

\theoremstyle{plain}

\theoremstyle{definition}

\theoremstyle{remark}

\crefname{assumption}{Assumption}{Assumptions}
\Crefname{assumption}{Assumption}{Assumptions}
\newcounter{algorithm}
\renewcommand{\thealgorithm}{\arabic{algorithm}}

\crefname{algorithm}{Algorithm}{Algorithms}
\Crefname{algorithm}{Algorithm}{Algorithms}

\newcommand{\R}{\mathbb{R}}

\DeclareMathOperator{\softmax}{softmax}

\newcommand{\cL}{\mathcal{L}}

\newcommand{\phisdf}{\boldsymbol{\phi}}
\newcommand{\wphy}{\mathbf{w}_{\text{phy}}}
\newcommand{\Ihist}{\mathbf{I}_{\text{hist}}}

\newcommand{\Ihat}{\hat{\mathbf{I}}}

\title{Physics-Guided Spatiotemporal State Space Modeling for Lookahead Molten Pool Segmentation in Laser Wire-Feed Welding}

\author{Sen Li, Haichao Cui\thanks{Corresponding authors: Haichao Cui and Fenggui Lu.}, Changhao Yin, Chendong Shao,\\
Yaqi Wang, Xinhua Tang, and Fenggui Lu\footnotemark[1]\\[0.25em]
\small Shanghai Key Laboratory of Materials Laser Processing and Modification,\\
\small School of Materials Science and Engineering, Shanghai Jiao Tong University,\\
\small Shanghai 200240, PR China}
\date{}

\begin{document}

\maketitle

\begin{abstract}
Real-time weld-pool perception is critical for closed-loop control in laser wire-feed welding, where sensing, computation, and actuator response introduce unavoidable delay.
This paper presents a physics-guided spatiotemporal state space network for lookahead weld-pool segmentation.
The model uses historical coaxial grayscale images, welding process parameters, and aligned wire-state electrical signals to predict the future semantic layout of three physically meaningful regions: keyhole, wire, and molten pool.
It combines a visual encoder, process- and sensor-conditioned feature normalization, patch-level temporal state space modeling, horizon-conditioned latent prediction, dense future feature prediction, and a motion-aware mask decoder.
Auxiliary signed-distance-function supervision, temporal consistency, feature distillation, and fine-grained keyhole losses further constrain the predicted geometry and local motion.
Experiments on a 43-sequence laser welding dataset show that the proposed WeldMamba reaches 74.63\% mIoU at a 500 ms lookahead.
Ablation studies further show that temporal history, patch-level state space modeling, and keyhole motion awareness are the main contributors to robust future segmentation.

\end{abstract}

\vspace{0.5em}
\noindent\textbf{Keywords:} Laser wire-feed welding; Weld pool segmentation;\
State space modelling; Lookahead prediction.
\vspace{0.5em}

\section{Introduction}
\label{sec:intro}

Laser welding has been extensively applied in automotive, aerospace, energy, shipbuilding, and high-end equipment manufacturing because it offers high energy density, low heat input, rapid welding speed, and a large weld depth-to-width ratio.
\citet{hong2017automotive_review} reviewed the prospects of laser welding in automotive manufacturing and emphasized its advantages in joining quality and production efficiency.
\citet{cai2020sensing_review} and \citet{lu2024laser_survey} further summarized sensing and machine-learning techniques for laser welding monitoring, showing that intelligent monitoring is becoming a key route toward stable industrial deployment.
Compared with conventional welding processes, laser welding provides stronger automation potential and more concentrated thermal action, but high-quality welded joints still require precise control of the process parameters and the transient molten pool state.
In laser wire-feed welding, stable fusion depends on the coupled evolution of the keyhole, filler wire, and molten pool.
Small changes in laser power, travel speed, wire feed rate, focal position, and beam oscillation mode can affect heat input and fluid flow, resulting in different molten pool states and ultimately influencing the quality of the welded joint.
Therefore, real-time perception of these regions is essential for controlling penetration depth, preventing defects, and ensuring weld quality.
However, closed-loop control cannot rely only on post-process inspection or current-frame recognition: sensing, computation, and actuator response all introduce delay, so the monitoring model should be able to predict the future welding state.

During laser welding, a large number of process signals are generated, among which optical and visual signals have received particular attention because they contain rich information, are relatively easy to acquire, and are directly interpretable for welding engineers.
\citet{kim2012coaxial_keyhole} used coaxial monitoring to observe keyhole behavior during Yb:YAG laser welding and showed that keyhole morphology provides direct evidence of the penetration process.
\citet{you2014highpower_monitoring} combined high-speed photography with image processing for high-power laser welding monitoring, demonstrating that temporal visual features can reflect process instability.
\citet{luo2015pool_boundary} extracted molten pool boundaries and measured pool width from keyhole fiber laser welding images, proving that molten pool geometry can be quantitatively obtained from optical images.
From the defect-formation perspective, \citet{ai2018keyhole_porosity} linked porosity generation to unstable keyhole dynamics, confirming that molten pool images are not ordinary visual textures but physical observations of heat transfer, vapor pressure, and fluid flow.
These studies established the physical basis for image-based monitoring.
Nevertheless, their visual descriptors are mainly designed for current-state observation and cannot directly provide the future semantic layout of the keyhole, filler wire, and molten pool required by a delayed closed-loop controller.

With the development of machine learning, researchers have increasingly used data-driven models to replace manually designed visual indicators.
\citet{zhao2019dbn_status} proposed a DBN-based online welding-status monitoring method and showed that learned features can describe the process state from sensing data.
\citet{zhang2020penetration_cnn} used a convolutional neural network for real-time penetration-state recognition in laser welding of tailor-rolled blanks, verifying the feasibility of direct image-to-state prediction.
\citet{zhang2020porosity_dl} developed a compact deep-learning model for porosity monitoring, and \citet{kim2021coaxial_monitoring} used synchronized coaxial observation to realize real-time full-penetration monitoring of laser keyhole welding.
To improve robustness beyond a single image stream, \citet{you2019multioptical_defects} detected thick-plate welding defects with multiple optical sensors, \citet{li2023cross_attention_fusion} introduced cross-attention fusion for multi-sensing penetration-state monitoring, and \citet{kang2023multisensor_depth} estimated weld penetration depth from deep models and multi-sensor signals.
These works show that deep learning and sensor fusion can improve welding status recognition and quality prediction.
However, their outputs are mostly penetration classes, defect decisions, scalar depths, or control-related quantities; they do not preserve the pixel-level geometry of the keyhole, filler wire, and molten pool, which limits their use when the controller needs spatially resolved future information.

Semantic segmentation provides a more direct description of molten pool morphology because it assigns a physical label to each image region.
\citet{cai2024surface_morphology} developed a lightweight attention-enhanced segmentation model for real-time weld surface morphology monitoring during laser keyhole welding, showing that dense visual prediction can satisfy online requirements.
Beyond laser welding, \citet{ma2024gmaw_lstm_unet} analyzed molten pool region constituents in GMAW and reconstructed dynamic regions with an LSTM U-Net structure.
\citet{zhang2023weld_area} predicted weld area in laser oscillation welding by combining image recognition and machine learning, further demonstrating the value of morphology-aware prediction for welding quality assessment.
However, most existing segmentation methods estimate the current frame or surface morphology, and the filler wire is rarely modeled jointly with the keyhole and molten pool.
For laser wire-feed welding, this is a critical limitation because the keyhole, wire-tip state, droplet-transfer behavior, and molten pool behavior are highly coupled.

Another important trend is the incorporation of welding parameters and physical priors into learning models.
\citet{li2026multitask_spatiotemporal} combined top molten pool images with welding parameters in a multi-task spatiotemporal network and achieved accurate prediction of penetration state, penetration depth, and weld cross-sectional morphology.
This result shows that welding parameters are not only auxiliary experimental records, but also useful explanatory variables for visual molten pool evolution.
\citet{li2026simphysnet} further introduced physics-informed feature learning for laser-welding penetration prediction and showed that physical priors can improve feature quality when annotations are limited.
These studies indicate that welding perception models should be conditioned by measurable process information rather than trained as image-only predictors.
However, their prediction targets remain penetration status, depth, or reconstructed cross-sectional morphology; they do not output the future pixel-level layout of the visible keyhole, filler wire, and molten pool.

Temporal modeling is also necessary because a single image cannot fully describe molten pool deformation, keyhole oscillation, and filler-wire transfer.
In laser keyhole welding, \citet{lu2020interaction_keyhole} used interaction-time-conditioned keyhole behaviors with an SVM classifier and showed that temporal keyhole descriptors improve penetration-status monitoring.
\citet{yu2022_dynamic_pool_cnn_lstm} used dynamic molten pool image sequences with a CNN-LSTM model for penetration monitoring and demonstrated that consecutive images provide more reliable information than isolated frames.
For laser-MIG hybrid welding, \citet{ye2022_top_vision_back_width} predicted weld back width from top-view visual sensing, while \citet{fan2024_informer_weld_widths} introduced an Informer-based model to predict weld widths from temporal process observations.
\citet{zhao2024_continuous_video} further studied dynamic penetration prediction through continuous video learning, indicating that welding state estimation benefits from models that treat the visual stream as a continuous process rather than a set of independent images.
These studies move welding monitoring from static image recognition toward temporal process understanding, but their outputs are still mainly penetration states or weld-width variables.
They do not produce future dense masks for the visible keyhole, filler wire, and molten pool.

Recent work on scanning and oscillating welding makes this limitation more evident.
\citet{ke2023_oscillating_modes} analyzed heat transfer and melt flow in keyhole, transition, and conduction modes during laser beam oscillating welding, showing that the apparent keyhole and molten pool behavior is tied to oscillation-induced thermal-fluid dynamics.
\citet{hong2024_af_fttsnet} designed a two-stream network for online robotic-welding quality monitoring, and \citet{hong2024_multigranularity_gta} used multigranularity spatiotemporal attentive representation learning for climbing GTAW, both confirming that temporal visual structures and multi-scale features are useful for industrial welding monitoring.
Most relevant to our setting, \citet{yan2025_lsw_crossvit} constructed a time-series dataset of keyhole and molten pool visual signals for laser scanning welding and proposed a dual-branch Transformer with self-attention and cross-attention to fuse the two dynamic regions.
Their model achieved 99.3\% testing accuracy for joint formation classification, and their ablation and attention visualization showed that temporal characteristics and fine-grained keyhole cues are decisive for recognizing scanning-welding states.
This line of work strongly supports the need to model local temporal dynamics in molten pool imagery.
However, it still treats keyhole and molten pool segmentation as an upstream representation step and makes a sequence-level state decision, whereas closed-loop laser wire-feed welding requires a future pixel-level layout that can be directly related to spatial control actions.

Visual prediction and sequence modeling provide tools for moving from current-image monitoring to delayed welding control.
\citet{liu2017video} estimated future video frames by learning pixel motion, showing that image histories can be used to infer future visual states.
\citet{ha2018world} and \citet{hafner2020dream} developed latent dynamics models in which future states are predicted in a compact representation space, providing a useful idea for control-oriented prediction.
More recent feature-prediction studies further show that future visual states can be estimated in latent space without reconstructing every pixel.
\citet{bardes2024vjepa} revisited video feature prediction, while \citet{kim2024cats} and \citet{wang2024timexer} showed that horizon-dependent queries and external variables can provide useful context for long-horizon forecasting.
\citet{fan2026prospect} also used learnable stream query tokens to predict next-step latent visual features.
For laser wire-feed welding, these ideas are useful because the controller acts after sensing, computation, and actuator delay; however, the prediction must preserve the local geometry of the keyhole, wire, and molten pool rather than only output a compact process state.
Efficient sequence modeling is therefore important.
\citet{dao2024mamba2} connected selective state space models with attention-like computation through structured state space duality, providing an efficient basis for long visual-temporal sequences.
In visual recognition, \citet{zhu2024vim}, \citet{liu2024vmamba}, and \citet{yang2024plainmamba} applied state space modeling to visual representation learning, and \citet{hesham2025tv3s} propagated semantic information across frames for dense video segmentation.
These works suggest that state space models are suitable for efficient temporal reasoning, but many temporal models compress the sequence into a global state.
Such compression may lose local keyhole and wire-tip motion; therefore, this paper uses patch-level temporal state modeling and shifted-window communication for future mask prediction.

Based on the above discussion, the key requirement for laser wire-feed welding monitoring is no longer only to recognize whether the present frame corresponds to a stable or defective state.
A practical monitoring model should describe the future spatial distribution of the keyhole, filler wire, and molten pool while using the process parameters and synchronized sensor signals that explain the observed welding dynamics.
To address this requirement, this paper proposes WeldMamba, a physics-guided spatiotemporal state space network for lookahead weld-pool segmentation.
The central idea is to inject welding process parameters and aligned wire-state electrical signals into the visual-temporal representation, so that the network can adapt its features and future prediction to the physical conditions that govern molten pool evolution.
The model combines a physics-conditioned encoder, a patch-level temporal state space network, dense future feature prediction, and a motion-aware future-mask decoder.

The contributions of this paper are summarized as follows:
\begin{enumerate}[leftmargin=*,itemsep=2pt]
    \item A \textbf{physics-conditioned multi-scale encoder} that adapts visual features using process parameters and current signals, while retaining high-resolution skip features for small weld structures.
    \item A \textbf{patch-level temporal state space network} with shifted-window communication and previous-mask markers, preserving local temporal dynamics without collapsing each frame to a single global vector.
    \item A \textbf{horizon-conditioned future prediction module} that estimates both a future latent and a dense future feature map, followed by a learned mask decoder with future-token cross-attention and keyhole-motion awareness.
    \item A \textbf{geometry- and dynamics-aware training objective} that combines supervised segmentation, SDF rendering, cross-condition feature alignment, DINOv3 feature distillation, fine-scale keyhole supervision, and motion regularization.
\end{enumerate}

The rest of this paper is organized as follows.
\Cref{sec:method} first defines the experimental data and lookahead segmentation task, and then presents the proposed architecture and training objective.
\Cref{sec:experiments} reports the experimental verification and discussion.
\Cref{sec:conclusion} concludes the paper and discusses limitations and future work.

\section{Materials and Methods}
\label{sec:method}

\subsection{Experimental Setup}
\label{sec:experimental_setup}

\begin{figure}[H]
\centering
\includegraphics[width=0.98\textwidth]{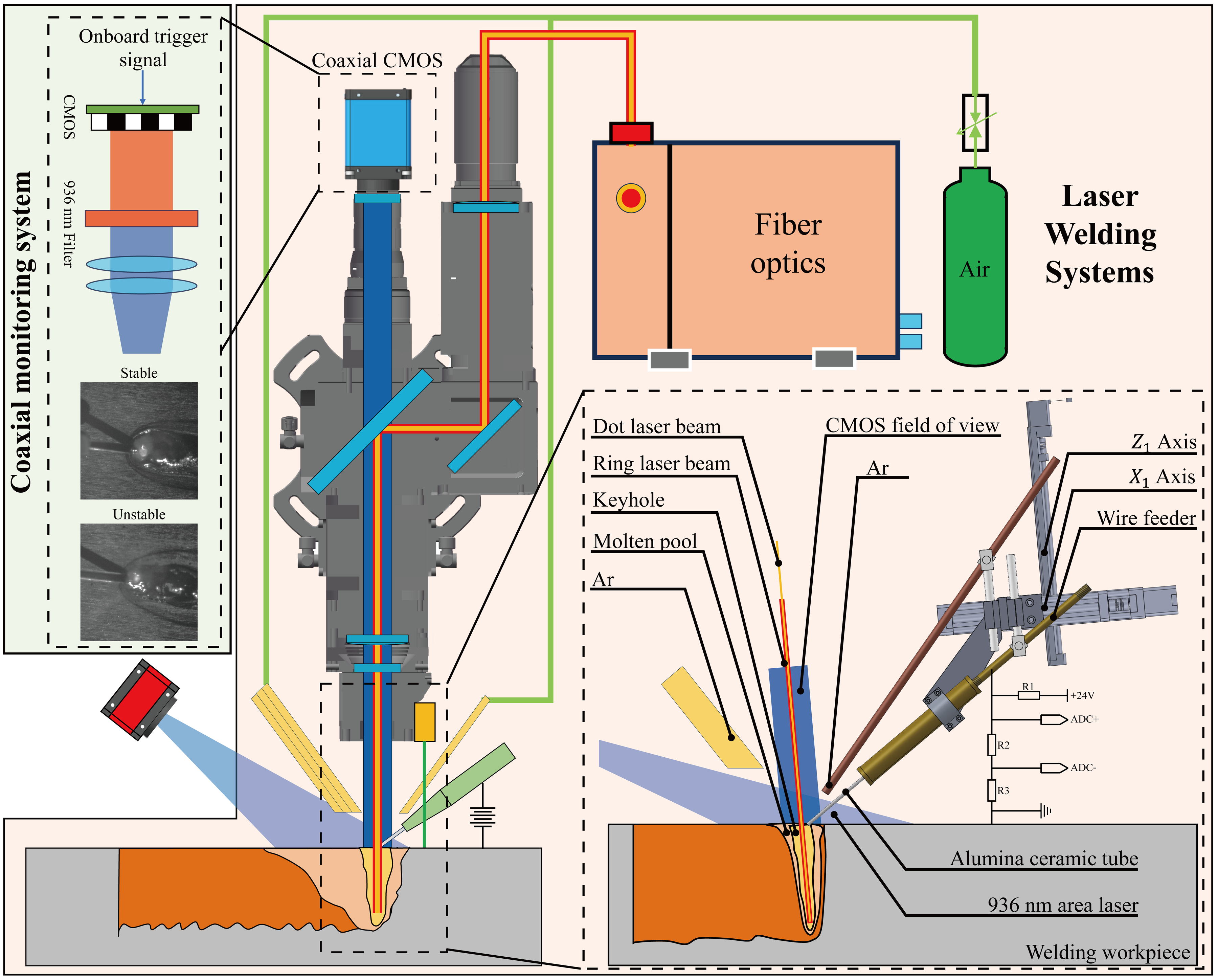}
\caption{Schematic of the laser wire-feed welding experimental system, including the laser module, coaxial imaging path, wire-feeding mechanism, shielding gas, workpiece fixture, and wire-state monitoring circuit.}
\label{fig:experimental_setup}
\end{figure}

The experimental system is used for image and signal acquisition during laser wire-feed welding, as shown in \Cref{fig:experimental_setup}.
It consists of five main parts: a laser welding module, a robotic motion-control system, an independently movable wire-feeding system, a coaxial image-acquisition module, and a wire-state monitoring module.
The laser welding module includes a fiber laser, a water-cooling unit, and a laser welding head.
The laser source is a YLS-20000 fiber laser with a spot-ring beam configuration and a maximum output power of 20\,kW.
For the annular ring beam, the beam parameter product is 15.377\,$\mathrm{mm}\cdot\mathrm{mrad}$, the focused spot diameter is 596.364\,$\mu$m, and the Rayleigh length is 26.25\,mm.
For the central spot beam, the beam parameter product is 3.385\,$\mathrm{mm}\cdot\mathrm{mrad}$, the focused spot diameter is 197.535\,$\mu$m, and the Rayleigh length is 2.88\,mm.

The coaxial image-acquisition module uses a monochrome CMOS camera to acquire top-view molten pool images, and a 936\,nm auxiliary light source is integrated into the optical path.
During image acquisition, the auxiliary light source triggers the CMOS camera to maintain stable illumination.
The imaging system uses a $1/4$-inch monochrome CMOS sensor and a narrow-bandpass filter centered at 936\,nm with a half-bandwidth of 30\,nm.
The camera exposure time is set to 4.3\,$\mu$s, with a gamma value of 6.4 and a black level of 1.2\%.
Before welding, the 936\,nm auxiliary light source and the focal position of the coaxial camera are adjusted to obtain clear, stable, and uniformly illuminated molten pool images.
During welding, the workpiece remains fixed, while a Yaskawa robotic arm drives the welding head along the prescribed trajectory.
The captured images are transmitted in real time to a connected PC for storage.
The PC controls the output power of the fiber laser, whereas process parameters such as welding speed and defocus distance are set through the robotic control system.

In preliminary experiments, it was difficult to determine from coaxial images alone whether the molten wire was undergoing droplet transfer or liquid-bridge transfer.
Therefore, a wire-state monitoring module was developed.
This module consists of three resistors, a constant-voltage power supply, and a voltage acquisition module.
The three resistors, together with the welding wire and the workpiece, form a voltage-divider circuit.
When the wire is in contact with the workpiece, current flows through $R_1$ and the wire to the workpiece, and the voltage across $R_2$ is therefore 0\,V.
When the wire is not in contact with the workpiece, current flows through $R_2$ and $R_3$ to the workpiece, and the voltage across $R_2$ is greater than 5\,V.
By recording the voltage variation across $R_2$, the transfer state of the molten wire can be reliably acquired during welding; the corresponding circuit connection is illustrated in the lower-right inset of \Cref{fig:experimental_setup}.

The base material is COST-E, and the specimen dimensions are 200\,mm $\times$ 80\,mm $\times$ 10\,mm (length $\times$ width $\times$ height).
The filler wire is ER90S-B9 with a diameter of 1.2\,mm.
The chemical compositions of the filler wire and base material are listed in \Cref{tab:chemical_composition}.

\begin{table}[t]
\caption{Chemical compositions of COST-E base material and ER90S-B9 filler wire (wt.\%)~\citep{gianfrancesco2016materials}.}
\label{tab:chemical_composition}
\centering
\small
\setlength{\tabcolsep}{4pt}
\begin{tabular}{lcccccccccc}
\toprule
Material & C & Mn & Si & Ni & Cr & Mo & W & Nb & V & Fe \\
\midrule
COST-E & 0.12 & 0.45 & 0.10 & 0.75 & 10.5 & 1.0 & 1.0 & 0.05 & 0.2 & Bal. \\
ER90S-B9 & 0.10 & 0.80 & 0.32 & 0.60 & 9.0 & 1.0 & -- & -- & 0.2 & Bal. \\
\bottomrule
\end{tabular}
\end{table}

Before welding, the specimens are thoroughly cleaned with anhydrous alcohol to remove contaminants that may affect weld-joint quality.
During welding, argon is used as the shielding gas to protect the weld region.
The argon is supplied through three parallel copper tubes located behind the weld seam and one copper tube located in front of the weld seam, with a flow rate of 15\,L/min and a purity of 99.99\%.
The welding parameters are summarized in \Cref{tab:welding_parameters}.
Different weld-joint states are obtained by varying the wire feeding speed $v_f$, spot laser power $P_d$, ring laser power $P_r$, laser oscillation width $W_o$, and oscillation frequency $f_o$, while the defocus amount $D_f$ and welding travel speed $v_w$ are fixed in the listed configuration.

\begin{table}[t]
\caption{Welding parameters used for data acquisition.}
\label{tab:welding_parameters}
\centering
\small
\setlength{\tabcolsep}{6pt}
\begin{tabular}{p{0.42\linewidth}p{0.48\linewidth}}
\toprule
Parameter & Value \\
\midrule
Spot laser power $P_d$ [kW] & 2.0, 2.25, 2.5, 2.75, 3.0 \\
Ring laser power $P_r$ [kW] & 0.0, 0.25, 0.5, 0.75, 1.0 \\
Wire feeding speed $v_f$ [m/min] & 0.8, 1.0, 1.2, 1.4, 1.6, 2.0 \\
Laser oscillation width $W_o$ [mm] & 0.0, 0.5, 1.0, 1.5 \\
Laser oscillation frequency $f_o$ [Hz] & 40, 60, 80, 100 \\
Defocus amount $D_f$ [mm] & +10.0 \\
Welding travel speed $v_w$ [mm/s] & 6.0 \\
\bottomrule
\end{tabular}
\end{table}

\subsection{Experimental Data and Problem Definition}
\label{sec:data}

Based on the above acquisition system, each welding sequence contains synchronized top-view molten pool images, process parameters, and wire-state electrical signals.
The visual stream records the keyhole, wire, and molten pool morphology, while the process parameters and voltage response describe the operating condition and wire-transfer state that drive these visual changes.
This setting follows the practical monitoring scenario in which images and process signals are collected simultaneously and must be interpreted together for closed-loop control.

The raw image stream is captured at 50\,fps with $640\times640$ grayscale frames.
Pixel-level annotations are provided for three foreground semantic categories: keyhole, wire, and molten pool.
\revone{The corresponding process vector contains seven normalized welding parameters: spot laser power, ring laser power, welding travel speed, wire feeding speed, defocus amount, laser-oscillation width, and laser-oscillation frequency.}
\revone{In the experiments reported here, welding travel speed and defocus amount are fixed by the acquisition plan; they are retained as part of the seven-parameter vector to keep a consistent interface across process recipes.}
The wire-state electrical signal is sampled at 10\,kHz and aligned to the image frames, so each training sample contains an image sequence and its synchronized electrical response signal.

\begin{figure}[H]
\centering
\includegraphics[width=0.98\textwidth]{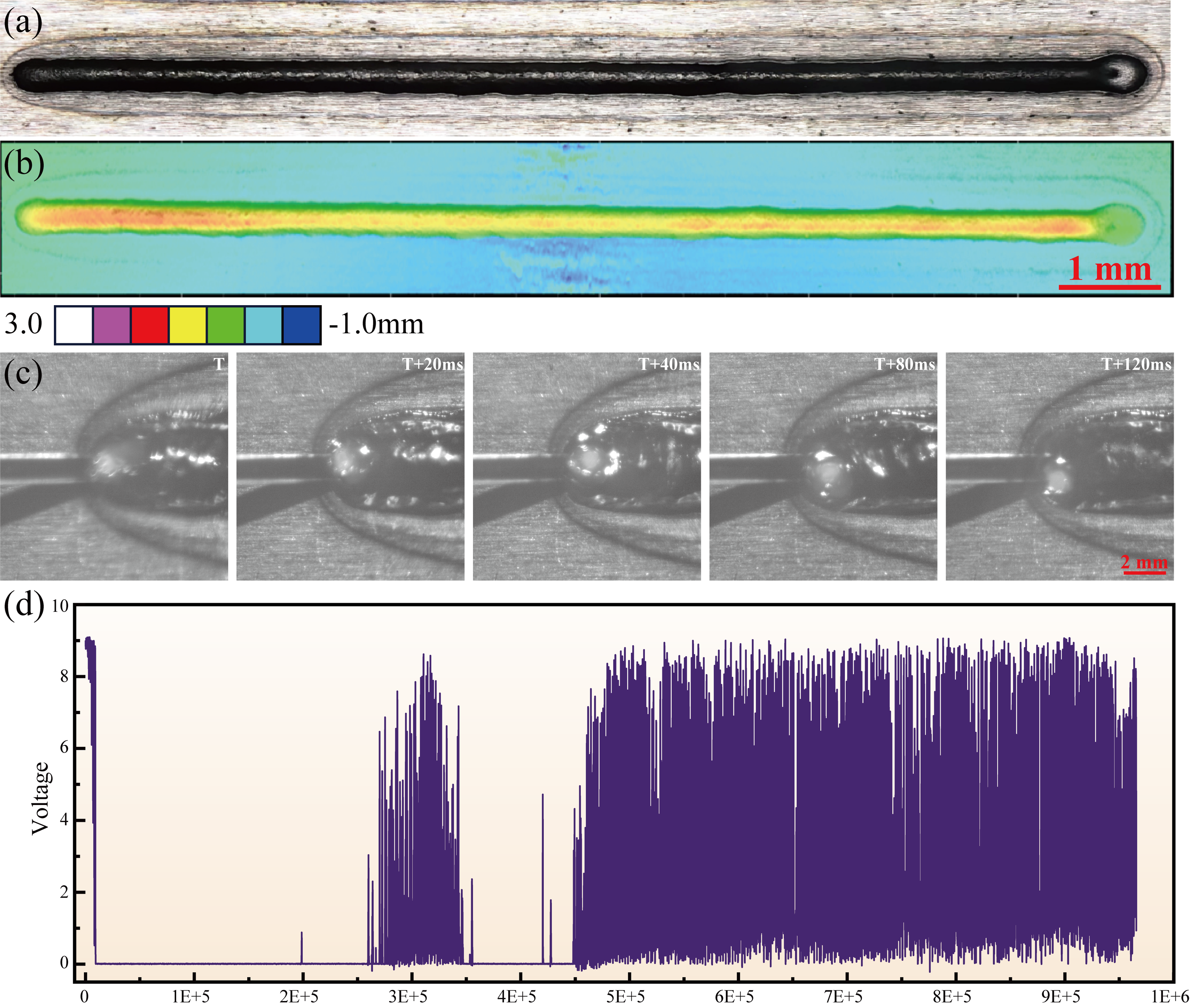}
\caption{Representative data acquired during an actual laser wire-feed welding process. The figure shows the post-weld bead appearance, a surface morphology visualization, a sequence of coaxial molten pool images at different times, and the aligned wire-state voltage signal.}
\label{fig:data_sequence}
\end{figure}

\begin{figure}[H]
\centering
\includegraphics[width=0.98\textwidth]{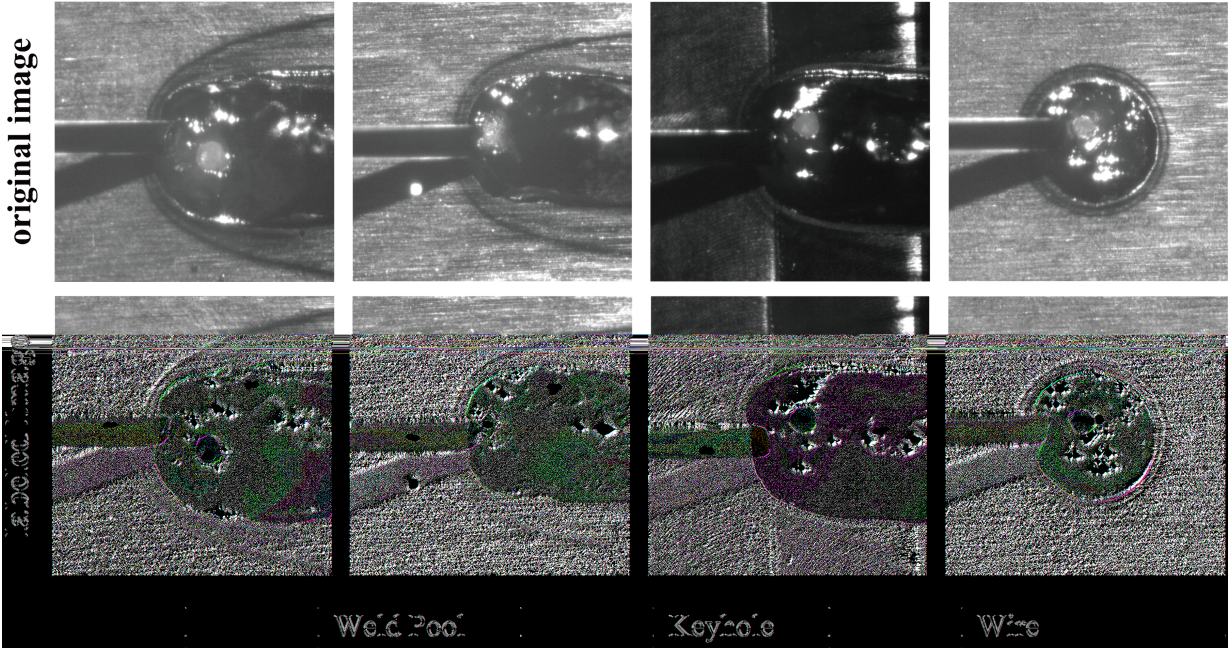}
\caption{Representative pixel-level annotations in the welding image dataset. The first row shows original coaxial grayscale images, and the second row overlays labels for the three target regions: molten pool, keyhole, and wire.}
\label{fig:annotation_examples}
\end{figure}

Representative acquisition examples are shown in \Cref{fig:data_sequence}.
\revone{The bead photograph, surface morphology visualization, coaxial image sequence, voltage curve, and annotation examples in \Cref{fig:data_sequence,fig:annotation_examples} are real acquisitions or labels from the collected welding dataset; the architecture figures are author-drawn schematics of the proposed model.}
The figure includes the weld-bead appearance, a surface morphology visualization, a time-ordered coaxial image sequence, and the aligned voltage response recorded during welding.
Representative pixel-level annotations are shown in \Cref{fig:annotation_examples}, where the molten pool, keyhole, and wire are labeled as three target regions.
These examples illustrate two practical properties of the dataset: the wire-transfer state is not always visually separable from the grayscale image alone, and the target regions differ strongly in scale, shape, and motion.
\revone{The pixel labels used in this work are stored as per-frame JSON annotations. Brush-style masks are decoded from run-length encodings, and legacy polygon annotations are rasterized before training. The current dataset release does not include a multi-annotator agreement study; boundary uncertainty around the small keyhole is therefore treated as a limitation rather than a separately quantified source of variance.}

The dataset contains 43 welding sequences acquired under different process conditions.
A sequence-level split is used to prevent adjacent frames from the same welding process from appearing separately in the training and validation sets.
We use an 80/20 split with a fixed random seed, resulting in 34 sequences for training and 9 sequences for validation.
\Cref{tab:dataset_composition} summarizes the resulting dataset composition.

\begin{table}[t]
\caption{Composition of the fully annotated welding image dataset under the adopted sequence split.}
\label{tab:dataset_composition}
\centering
\small
\setlength{\tabcolsep}{10pt}
\begin{tabular}{lrr}
\toprule
Split & Sequences & Annotated frames \\
\midrule
Train & 34 & 32,358 \\
Validation & 9 & 8,361 \\
Total & 43 & 40,719 \\
\bottomrule
\end{tabular}
\end{table}

Given the historical image sequence $\Ihist$, the process parameter vector $\mathbf{p}$, and the aligned wire-state electrical signal $\mathbf{c}$, the task is to predict the semantic mask $K$ frames ahead.
Formally, $\Ihist=\{I_{t_1},\ldots,I_{t_T}\}\in\R^{T\times1\times H\times W}$ denotes the grayscale history, $\mathbf{p}\in\R^7$ denotes the normalized welding-process vector, and $\mathbf{c}$ denotes the aligned wire-state electrical signal over the same history window.
The model predicts a future multi-class mask $\mathbf{y}_{t+K}\in\R^{C\times H\times W}$ and signed distance functions $\boldsymbol{\phi}_{\mathrm{sdf}}\in\R^{C\times H\times W}$, where $C=3$ corresponds to keyhole, wire, and molten pool.
The three target regions have different physical meanings: the keyhole reflects the laser-material interaction and is small and rapidly moving; the wire represents filler-wire transfer and may be partially occluded; the molten pool describes the overall weld morphology.
These regions are not merely visual categories but the observable footprints of the mechanisms that generate defects.
The relative position and stability of the wire tip within the molten pool largely determine whether the filler metal wets the surrounding solid surface: when the relative position between the wire tip and the keyhole changes, the droplet cannot bridge to the appropriate location, which further alters weld morphology and is a primary cause of incomplete-fusion and porosity defects.
In addition, the wire tip is disturbed by the laser plume, whose volume, intensity, and position fluctuate with the keyhole and impinge on the wire; therefore, the spatial relationship among the keyhole, plume-affected wire, and molten pool directly reflects impending instability.
Accurate segmentation of these three regions is therefore essential for estimating the welding-process state because it precisely localizes the wire--molten pool coupling and keyhole behavior from which such defects originate.
We report mIoU over the three target classes, together with class-wise IoU for keyhole, wire, and molten pool.

Let $\hat{M}_c$ and $M_c$ denote the predicted and ground-truth binary masks for class $c$, respectively.
The class-wise IoU is defined as
\begin{equation}
    \mathrm{IoU}_c =
    \frac{|\hat{M}_c \cap M_c|+\epsilon}
    {|\hat{M}_c \cup M_c|+\epsilon},
\end{equation}
where $\epsilon$ is a small constant for numerical stability.
For a class set $\mathcal{S}$, the mean IoU is
\begin{equation}
    \mathrm{mIoU}(\mathcal{S}) =
    \frac{1}{|\mathcal{S}|}\sum_{c\in\mathcal{S}}\mathrm{IoU}_c .
\end{equation}
The reported mean metric uses $\mathcal{S}=\{\mathrm{keyhole},\mathrm{wire},\mathrm{molten\ pool}\}$.

\subsection{Details of the Network}
\label{sec:method_overview}

\begin{figure}[H]
\centering
\includegraphics[width=0.95\textwidth]{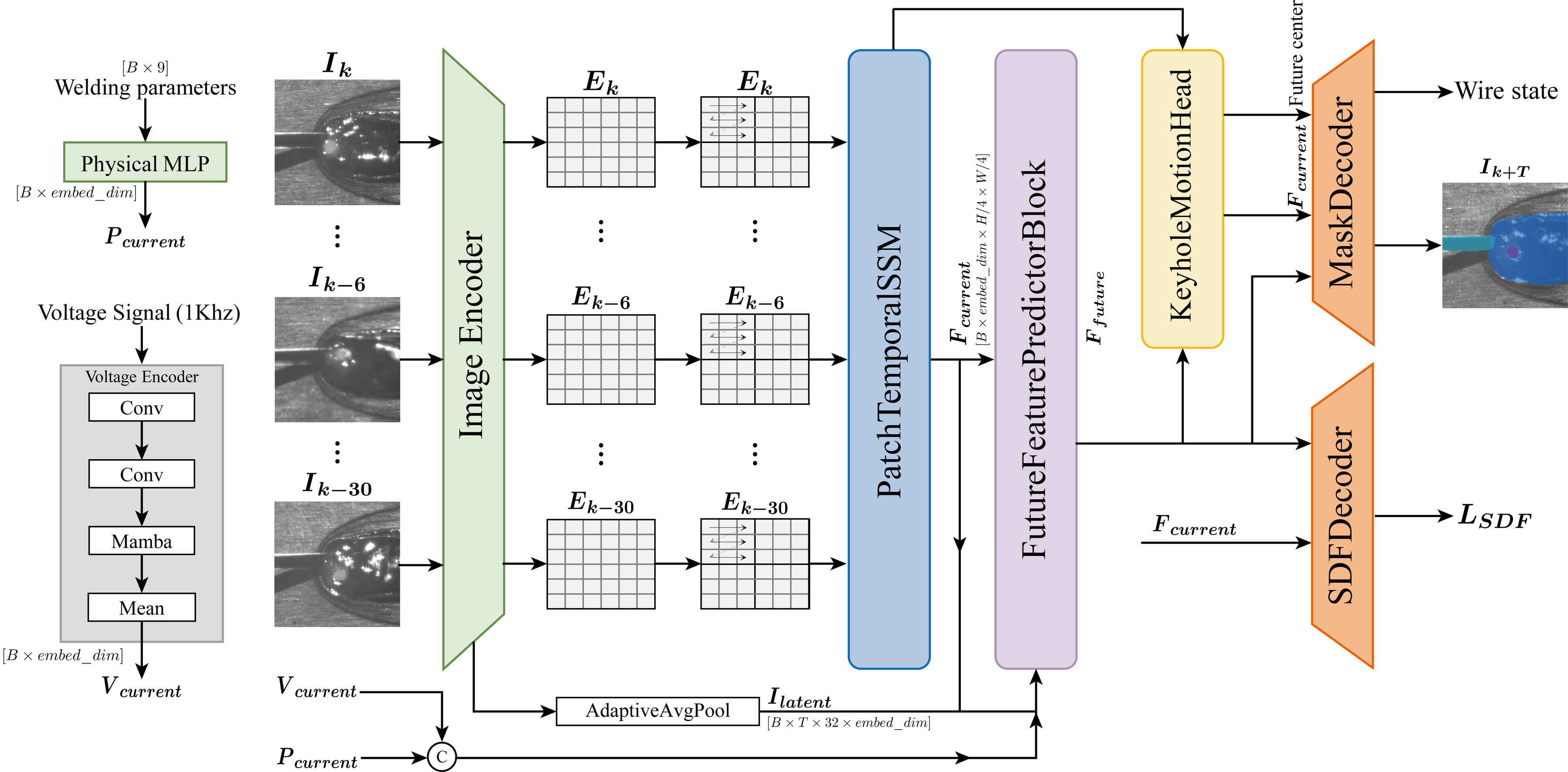}
\caption{\revone{Overview of WeldMamba.} Six historical frames, process parameters, and wire-state electrical signals are fused for lookahead segmentation of keyhole, wire, and molten pool. Auxiliary SDF rendering, fine-scale keyhole supervision, and feature distillation are used only during training.}
\label{fig:architecture}
\end{figure}

\revone{We refer to the proposed process- and sensor-conditioned spatiotemporal state space network as WeldMamba.}
The model uses a short-term history of welding images, process parameters, and the aligned wire-state electrical signal to predict the semantic mask and signed distance functions (SDFs) after a future time interval.
\revone{The architecture combines a pretrained MiT-B1 image encoder, process-conditioned feature normalization, a patch-level temporal state space network, dense future feature prediction, and a motion-aware mask decoding mechanism.}
The overall architecture is shown in \Cref{fig:architecture}.

\subsection{Physics and Sensor Conditioning}
\label{sec:physics_embed}

The appearance of a coaxial molten pool image is not arbitrary image texture.
Spot and ring laser powers, travel speed, wire feeding speed, defocus, and beam oscillation jointly determine heat input, recoil pressure, melt flow, and coupling between the wire and molten pool; these factors directly change keyhole size, wire melting behavior, and molten pool boundary shape.
The aligned wire-state electrical signal provides another source of process information because droplet transfer and liquid-bridge transfer can be difficult to distinguish from grayscale images alone.
\revone{We therefore form a conditioning vector $\wphy \in \R^d$ from two branches.}
The two branches are designed to encode complementary information at different temporal scales.
The process-parameter branch represents the relatively stable operating condition set before and during welding, whereas the wire-state electrical branch summarizes rapid transfer events observed within the recent history window.
After the two embeddings described below are projected to the same dimensionality, they are concatenated so that the visual encoder receives both the nominal welding setting and the short-term sensor response.
The final conditioning vector is $\wphy=[\mathbf{e}_p;\mathbf{e}_c]\in\R^d$, where $\mathbf{e}_p\in\R^{d/2}$ is the process-parameter embedding and $\mathbf{e}_c\in\R^{d/2}$ is the wire-state electrical embedding, each defined below.

\paragraph{Process parameter branch.}
The normalized process vector $\mathbf{p}\in\R^7$ is mapped to a $d/2$-dimensional embedding:
\begin{equation}
    \mathbf{e}_p = \mathrm{MLP}(\mathbf{p}) \in \R^{d/2},
\end{equation}
where the MLP uses hidden widths 32 and 64 with GELU activations.

\paragraph{Wire-state electrical branch.}
The wire-state electrical input is flattened over the history window and encoded by two strided 1D convolutions followed by a Mamba-2 layer:
\begin{equation}
    \mathbf{e}_c =
    \mathrm{MeanPool}\!\left(\mathrm{Mamba2}
    \left(\mathrm{Conv1D}_{1\rightarrow16\rightarrow d/2}(\mathbf{c})\right)\right)
    \in \R^{d/2}.
\end{equation}
If either input is unavailable, the corresponding half of the vector is filled with zeros, so the same network can be evaluated under partial sensor availability.

\subsection{Physics-Conditioned Image Encoder}
\label{sec:encoder}

The image encoder must preserve both small transfer-related structures and large pool morphology in single-channel coaxial images.
The keyhole requires high-resolution local features, whereas the molten pool outline and the wire outline require broader contextual features.
We therefore use a SegFormer MiT-B1 backbone~\cite{xie2021segformer} initialized from pretrained weights and adapted to single-channel welding imagery.
A lightweight CNN stem supplies high-resolution skip features:
\begin{align}
    \mathbf{f}_0 &= \mathrm{Stem}(I_t) \in \R^{32 \times H \times W},\\
    \mathbf{f}_1 &= \mathrm{Down}_1(\mathbf{f}_0) \in \R^{64 \times H/2 \times W/2}.
\end{align}
The MiT-B1 encoder produces four hierarchical feature maps at strides 4, 8, 16, and 32.
Each stage is projected to $d$ channels, resized to the stride-4 grid, concatenated, and fused:
\begin{equation}
    \mathbf{f}_2 =
    \mathrm{Fuse}\left(
    [P_1(\mathbf{s}_1), \mathrm{Up}(P_2(\mathbf{s}_2)),
    \mathrm{Up}(P_3(\mathbf{s}_3)), \mathrm{Up}(P_4(\mathbf{s}_4))]
    \right)
    \in \R^{d \times H/4 \times W/4}.
\end{equation}
We also retain a stride-8 feature $\mathbf{f}_3\in\R^{d\times H/8\times W/8}$ for cross-level fusion.

\paragraph{Physics-aware adaptive normalization.}
\revone{The fused stride-4 and stride-8 features are modulated by the conditioning vector through adaptive normalization:}
\begin{equation}
    \mathrm{CondNorm}(\mathbf{x},\wphy)
    =
    \gamma(\wphy)\odot
    \frac{\mathbf{x}-\mu(\mathbf{x})}{\sigma(\mathbf{x})+\epsilon}
    +\beta(\wphy),
\end{equation}
where $[\gamma;\beta]$ is produced by a linear layer from $\wphy$.
This allows the same local image contrast to be interpreted relative to the current welding condition, which is important because changes in power, wire feeding, defocus, and oscillation can shift pool brightness and shape without changing the semantic meaning of the region.

\paragraph{Cross-level feature fusion.}
Before temporal modeling, $\mathbf{f}_0,\mathbf{f}_1,\mathbf{f}_2,\mathbf{f}_3$ are projected to $d$ channels, resized to the stride-4 grid, and fused by a multi-scale pyramid:
\begin{equation}
    \mathbf{g}_t =
    \mathrm{MLP}_{\mathrm{fuse}}
    \left([\mathrm{Up}(Q_0\mathbf{f}_0),
    \mathrm{Up}(Q_1\mathbf{f}_1), Q_2\mathbf{f}_2,
    \mathrm{Up}(Q_3\mathbf{f}_3)]\right)
    + \mathbf{f}_2 .
\end{equation}
The sequence $\{\mathbf{g}_{t_i}\}_{i=1}^{T}$ is the dense input to the temporal module.
This fused representation keeps the fine contours of the keyhole and wire while retaining the wider molten pool context needed to infer heat accumulation and pool deformation.

\subsection{History-Aware Patch Temporal SSM}
\label{sec:temporal}

Motivated by temporally shared state-space processing in dense video segmentation, we construct a patch-level temporal state space network with residual Mamba blocks and shifted-window communication.
This choice follows the local nature of welding dynamics: keyhole displacement, wire-tip motion, and molten pool boundary deformation occur in neighboring image regions, but they evolve at different speeds.
The module therefore operates on dense feature maps rather than first compressing each frame to a single vector, which is important for small structures such as the keyhole and wire tip.

\paragraph{Previous-mask marker.}
Before temporal mixing, the model predicts an auxiliary soft mask for each historical feature map using a $1\times1$ head.
For frame $i>1$, the previous mask is converted to a learned class embedding map and added to the current dense feature:
\begin{equation}
    \bar{\mathbf{g}}_{t_i}
    =
    \mathbf{g}_{t_i}
    +
    \sum_{c=1}^{C}
    \mathrm{Down}(\mathbf{m}_{t_{i-1},c})\mathbf{a}_c ,
\end{equation}
where $\mathbf{a}_c\in\R^d$ is the learned embedding for class $c$.
This marker gives the temporal SSM explicit access to where semantic regions were in the recent past.
In welding videos this is useful because specular highlights, plume disturbance, and partial wire occlusion can make appearance-only correspondence unreliable.

\begin{figure}[H]
\centering
\includegraphics[width=0.62\textwidth]{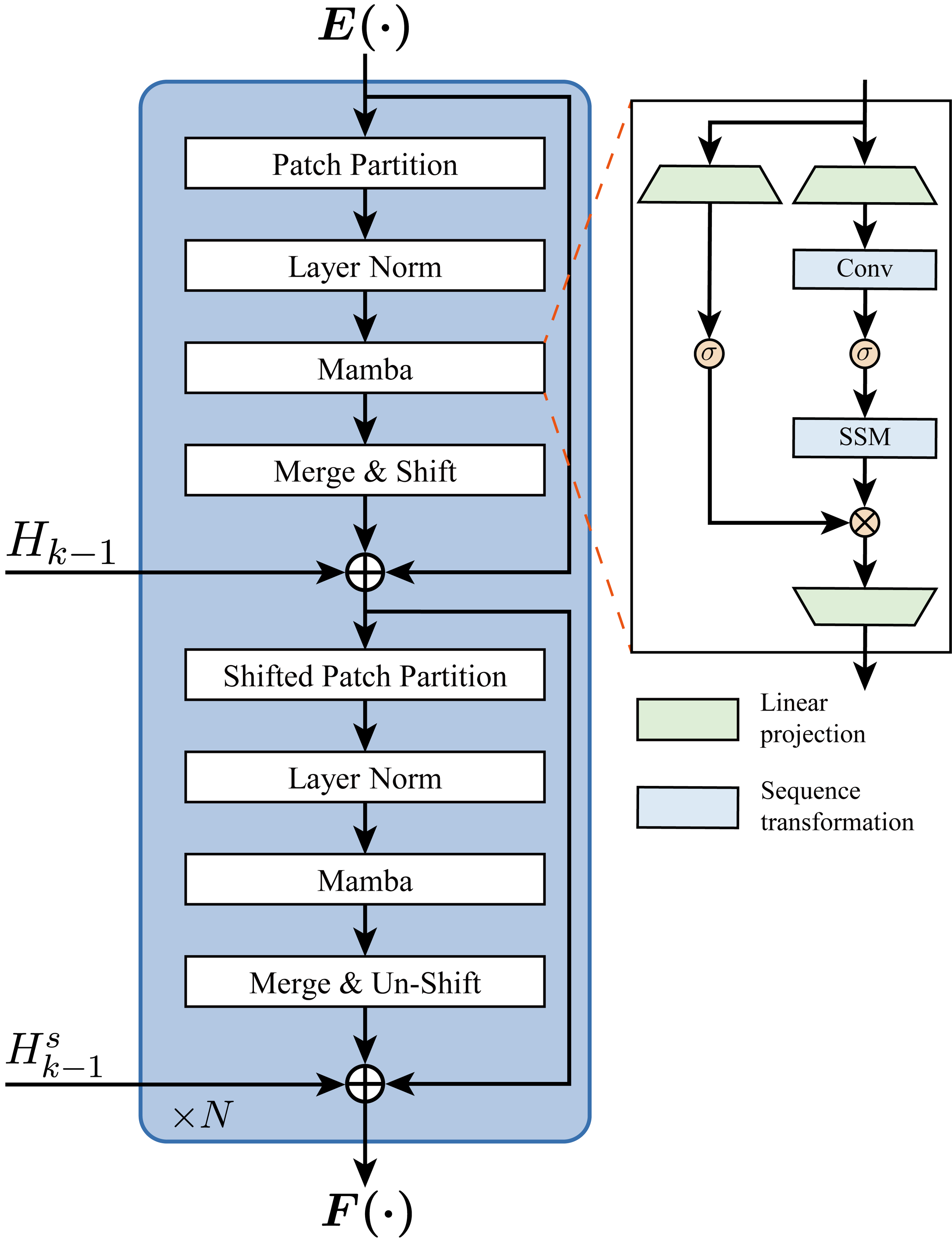}
\caption{Architecture of the PatchTemporalSSM block. Each stage pairs regular- and shifted-window Mamba-2 passes with residual injection of the previous hidden state. The expanded view shows the internal gating: two linear projections route the input through a depthwise convolution and an SSM before being re-combined by a sigmoid gate.}
\label{fig:patch_temporal_ssm}
\end{figure}

\paragraph{Shifted-window state space mixing.}
Each feature sequence is resized to a half-resolution grid whose dimensions are divisible by the window size.
With window size $w=20$ and shift $s=10$, the features are partitioned into windows and flattened over time and space:
\begin{equation}
    \mathbf{X}_{\mathrm{win}}
    \in
    \R^{(B n_w)\times (T w^2)\times d}.
\end{equation}
Four Mamba-2 layers process these window sequences.
Even layers use regular windows, while odd layers cyclically shift the grid by $s$ pixels before partitioning:
\begin{equation}
    \mathbf{X}^{(\ell+1)}
    =
    \begin{cases}
    \mathrm{Mamba2}(\mathrm{WinPart}(\mathbf{X}^{(\ell)})), & \ell \text{ even},\\
    \mathrm{Mamba2}(\mathrm{WinPart}(\mathrm{Shift}_{s}(\mathbf{X}^{(\ell)}))), & \ell \text{ odd}.
    \end{cases}
\end{equation}
Each block normalizes the window tokens, applies a Mamba-2 update, and adds the result back to the residual stream carried across layers.
The shifted windows allow local welding events to communicate across neighboring regions, which is needed when the keyhole drifts, the wire tip enters the pool, or the pool boundary expands beyond one patch.
After unpartitioning and upsampling to the stride-4 grid, the last temporal feature $\tilde{\mathbf{g}}_t$ is used as the current dense temporal state.
The detailed architecture of the PatchTemporalSSM block is shown in \Cref{fig:patch_temporal_ssm}.

\paragraph{Current temporal latent.}
The temporally mixed feature is globally pooled and projected to a current latent by a learned linear map $P_z$:
\begin{equation}
    \mathbf{z}_{t}= P_z\!\left(\mathrm{AvgPool}(\tilde{\mathbf{g}}_{t})\right)\in\R^d .
\end{equation}
For the future predictor, the model additionally pools each historical dense feature into a $4\times8$ grid of latent tokens, retaining coarse spatial layout rather than using only one global token per frame.

\subsection{Horizon-Conditioned Future Prediction}
\label{sec:future}

The final architecture does not rely only on repeated linear latent rollout.
For welding control, the requested horizon corresponds to the time interval that must cover sensing, computation, and actuator response delay.
This delay is not negligible in a real welding cell: image transfer and inference consume several frame intervals, and the manipulated variables themselves respond on different time scales.
Laser power is a fast actuator that directly modulates keyhole dynamics, yet a commanded power step still settles only after a finite transient on the order of tens of milliseconds, whereas the feed and wire-delivery mechanics react more slowly through their drive axes.
The slowest link in this chain sets the achievable closed-loop bandwidth, so a controller can only act on a process state that lies roughly one such delay in the future; a segmentation produced for the current frame is already stale by the time the laser or wire responds, and acting on it would chase a keyhole and molten pool that have since moved.
This matters because the quantities a controller must keep within bounds---keyhole depth and stability, molten pool size, and coupling between the wire and molten pool---can drift quickly, and an undetected keyhole collapse manifests as spatter and an abrupt penetration change.
At the 50\,fps acquisition rate of our system, a $K$-frame horizon corresponds to $K/50$ seconds of physical look-ahead, so the horizon can be selected according to the sensing-to-actuation latency of the control loop.
The network therefore predicts both a future global latent and a future dense feature map conditioned on the requested horizon, so that the output remains a spatial mask rather than a scalar state forecast.

\paragraph{Future latent query.}
A learned horizon embedding associated with the selected prediction horizon $K$ initializes a future query token.
The query is anchored by the current latent $\mathbf{z}_t$ and cross-attends to the sequence of historical latent tokens:
\begin{equation}
    \mathbf{q}_K^{(0)} = \mathbf{z}_{t} + \mathbf{e}_{K}, \qquad
    \mathbf{q}_K^{(r+1)}
    =
    \mathrm{AttnBlock}(\mathbf{q}_K^{(r)}, \mathbf{Z}_{\mathrm{hist}}).
\end{equation}
The predicted future latent is a residual update:
\begin{equation}
    \mathbf{z}_{t+K}=\mathbf{z}_t+\Delta(\mathbf{q}_K).
\end{equation}
The latent predictor itself does not require an additional wire-state token at this stage; the electrical response has already entered through the process-conditioned encoder and temporal SSM.

\begin{figure}[H]
\centering
\includegraphics[width=0.72\textwidth]{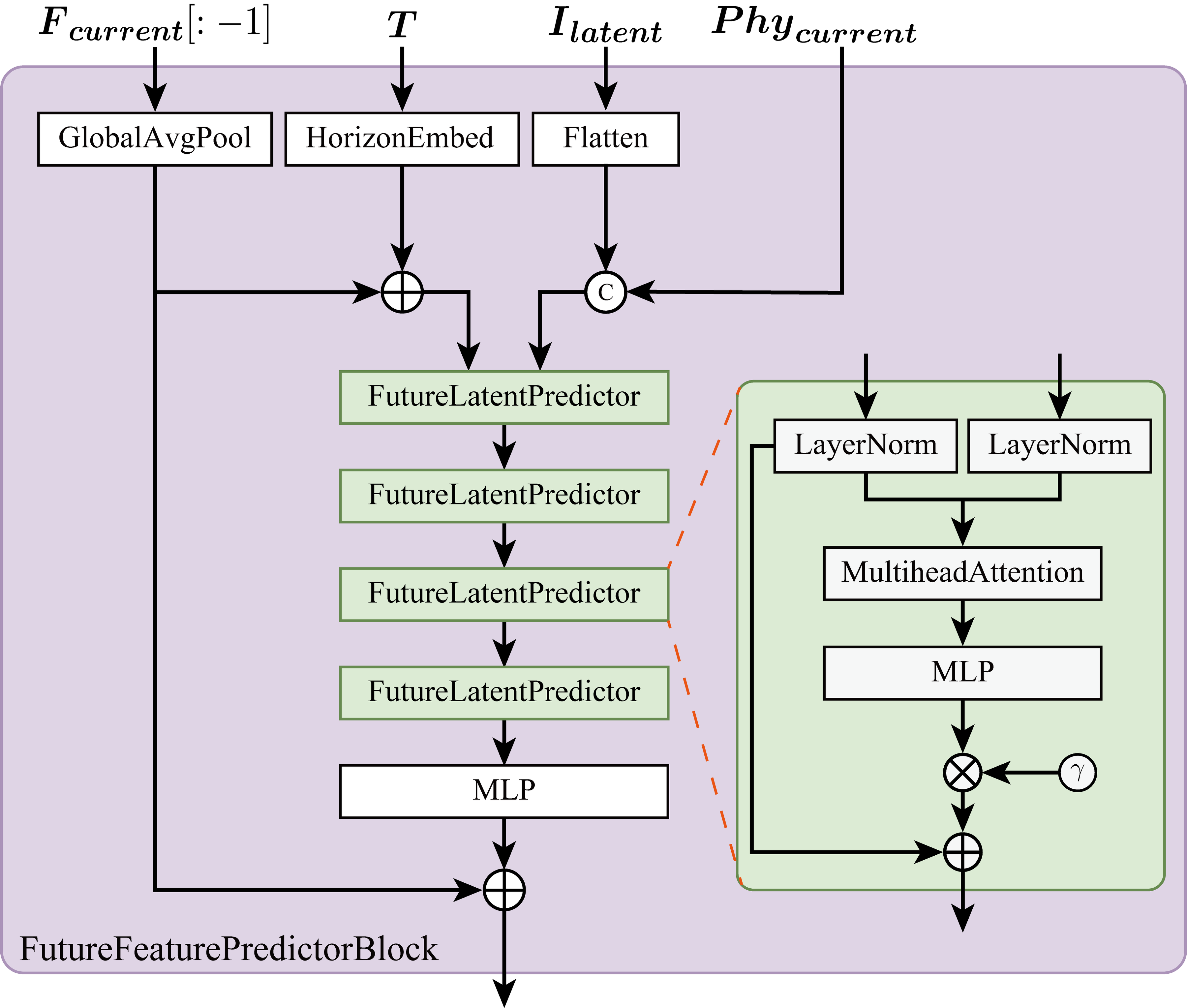}
\caption{Architecture of the FutureFeaturePredictorBlock. The historical feature summary, horizon embedding, image latent $I_{\mathrm{latent}}$, and conditioning vector $\mathrm{Cond}_{\mathrm{current}}$ are fused by a stack of FutureLatentPredictor attention blocks. The expanded view shows each block: two LayerNorm-gated cross-attention streams are combined with a learnable scale $\gamma$ and summed with a skip connection.}
\label{fig:future_feature_predictor}
\end{figure}

\paragraph{Dense future feature predictor.}
To avoid decoding the future mask from a global vector alone, we use a lightweight dense predictor.
This is important in laser wire-feed welding because future control-relevant information is spatial: the keyhole center, wire-entry position, and molten pool boundary must remain localized in the predicted feature map.
The last six dense history features are pooled to a compact $12\times12$ token grid.
A horizon-conditioned future query grid is concatenated with the history tokens, and alternating temporal and spatial attention layers complete the future grid.
The completed token grid is upsampled and refined together with four convolutional context terms:
\begin{equation}
    \mathbf{u}_K =
    \mathrm{TokenComplete}(\bar{\mathbf{g}}_{t-T+1:t}, \tilde{\mathbf{g}}_t, K),
\end{equation}
\begin{equation}
    \mathbf{f}^{\mathrm{future}}_2
    =
    \tilde{\mathbf{g}}_t
    +
    \rho\,
    \Delta_{\mathrm{dense}}\!\left(
    [\tilde{\mathbf{g}}_t,\mathbf{g}_t,
    \mathrm{Mean}(\bar{\mathbf{g}}),\mathbf{g}_t-\mathbf{g}_{t-T+1},
    \mathbf{u}_K]\right),
\end{equation}
where $\rho$ is a learnable residual scale.
During training, this module also emits auxiliary fine-resolution features and a mask hint, but only $\mathbf{f}^{\mathrm{future}}_2$ is required for deployment.
The internal structure of the FutureFeaturePredictorBlock used in the token-completion step is illustrated in \Cref{fig:future_feature_predictor}.

\subsection{Future Mask, SDF, and Motion-Aware Decoding}
\label{sec:decoder}

The network has two complementary output paths matched to welding monitoring requirements.
The main deployed output is a future semantic mask decoded directly from the future dense feature, because closed-loop decisions require the spatial layout of the keyhole, wire, and molten pool.
The auxiliary SDF path provides level-set supervision and differentiable rendering losses during training, encouraging smooth and physically plausible region boundaries for the continuous molten pool surface.
A signed distance field is a natural representation here because the molten pool is bounded by a smooth solid--liquid interface whose curvature reflects surface tension and the local thermal gradient; rendering this level set back to image space and matching it against the observed future frame supplies a geometry-aware signal that discourages frayed or topologically broken pool boundaries, which a pure per-pixel mask loss does not penalize.

\begin{figure}[H]
\centering
\includegraphics[width=0.88\textwidth]{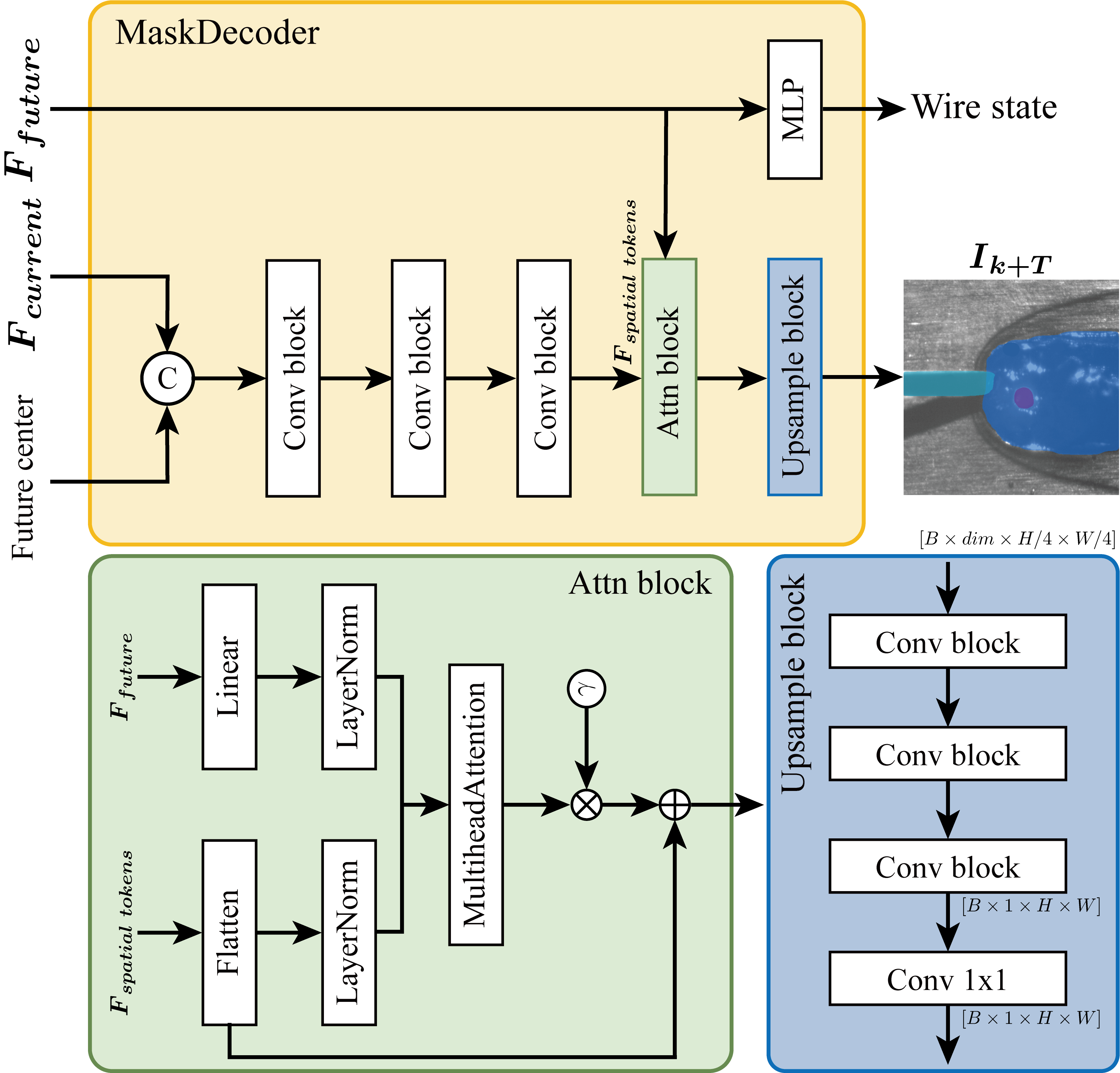}
\caption{Architecture of the future mask decoder. Dense future and current features are refined by convolutional blocks, cross-attention, and upsampling to produce the full-resolution lookahead mask.}
\label{fig:mask_decoder}
\end{figure}

\paragraph{Future mask decoder.}
The future dense feature is affine-modulated by $\mathbf{z}_{t+K}$, refined by local convolutions, and passed through a future-token cross-attention block:
\begin{equation}
    \mathbf{r}_K =
    \mathrm{XAttn}\left(
    \mathrm{Conv}\left(
    (1+\gamma(\mathbf{z}_{t+K}))\odot \mathbf{f}^{\mathrm{future}}_2
    + \beta(\mathbf{z}_{t+K})
    \right),
    \mathbf{z}_{t+K}
    \right).
\end{equation}
The mask decoder uses a local-MLP prediction head and a learned two-stage pixel-shuffle upsampler to produce full-resolution logits:
\begin{equation}
    \mathbf{y}_{t+K}=\mathrm{MaskHead}(\mathbf{r}_K)\in\R^{C\times H\times W}.
\end{equation}
This decoder deliberately avoids shallow last-frame skip connections, reducing the tendency to copy the current frame when predicting the future.
The detailed architecture of the mask decoder is shown in \Cref{fig:mask_decoder}.

\begin{figure}[H]
\centering
\includegraphics[width=0.40\textwidth]{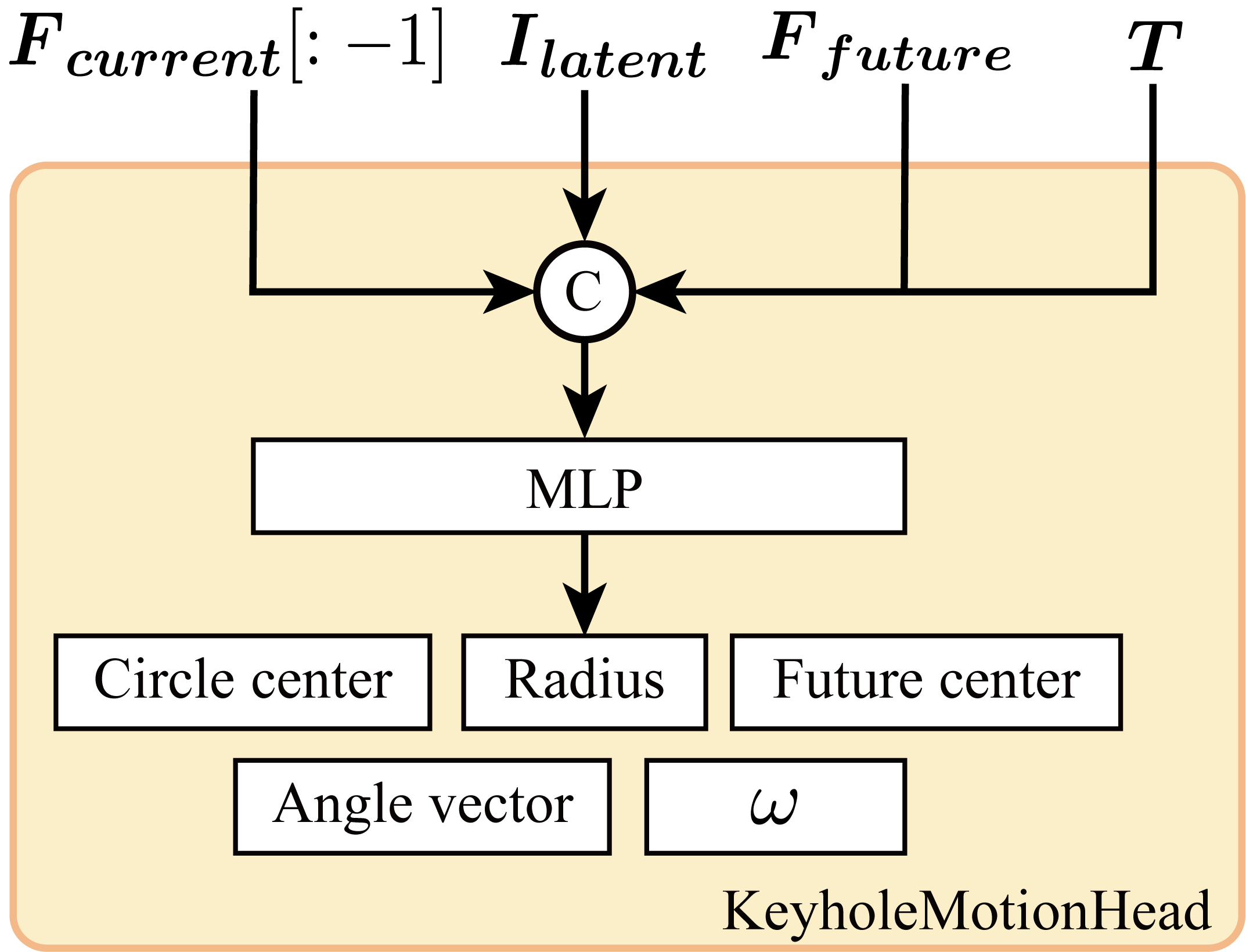}
\caption{Architecture of the KeyholeMotionHead, which predicts the circular-motion descriptor and future keyhole center from historical, dense future, and horizon-conditioned features.}
\label{fig:keyhole_motion_head}
\end{figure}

\paragraph{Keyhole motion awareness.}
Because the keyhole is small and moves rapidly under beam oscillation, recoil pressure, and local interaction between the wire and molten pool, we add a sample-level motion head.
The circular form of this descriptor is grounded in the physics of oscillating-beam laser welding: when the beam follows a circular oscillation path, the keyhole does not drift randomly but revolves around a rotation center, and the resulting stirring effect intensifies with oscillation frequency and amplitude, dragging the surrounding liquid metal along an approximately circular trajectory.
A descriptor that explicitly carries a center, a radius, and an angular velocity therefore matches the dominant motion mode of the keyhole far better than an unconstrained displacement vector, and it remains meaningful even when the keyhole is momentarily occluded by spatter or plume.
It attends from a horizon-conditioned query to historical latent tokens and dense temporal tokens, and predicts a circular-motion descriptor: center, radius, angular direction, angular velocity, confidence, and future center.
The predicted future center forms a Gaussian motion prior for the future-mask decoder, and a motion cross-attention module biases dense features toward the anticipated keyhole region instead of relying only on the last visible position.
The structure of the KeyholeMotionHead is shown in \Cref{fig:keyhole_motion_head}.

\begin{figure}[H]
\centering
\includegraphics[width=0.72\textwidth]{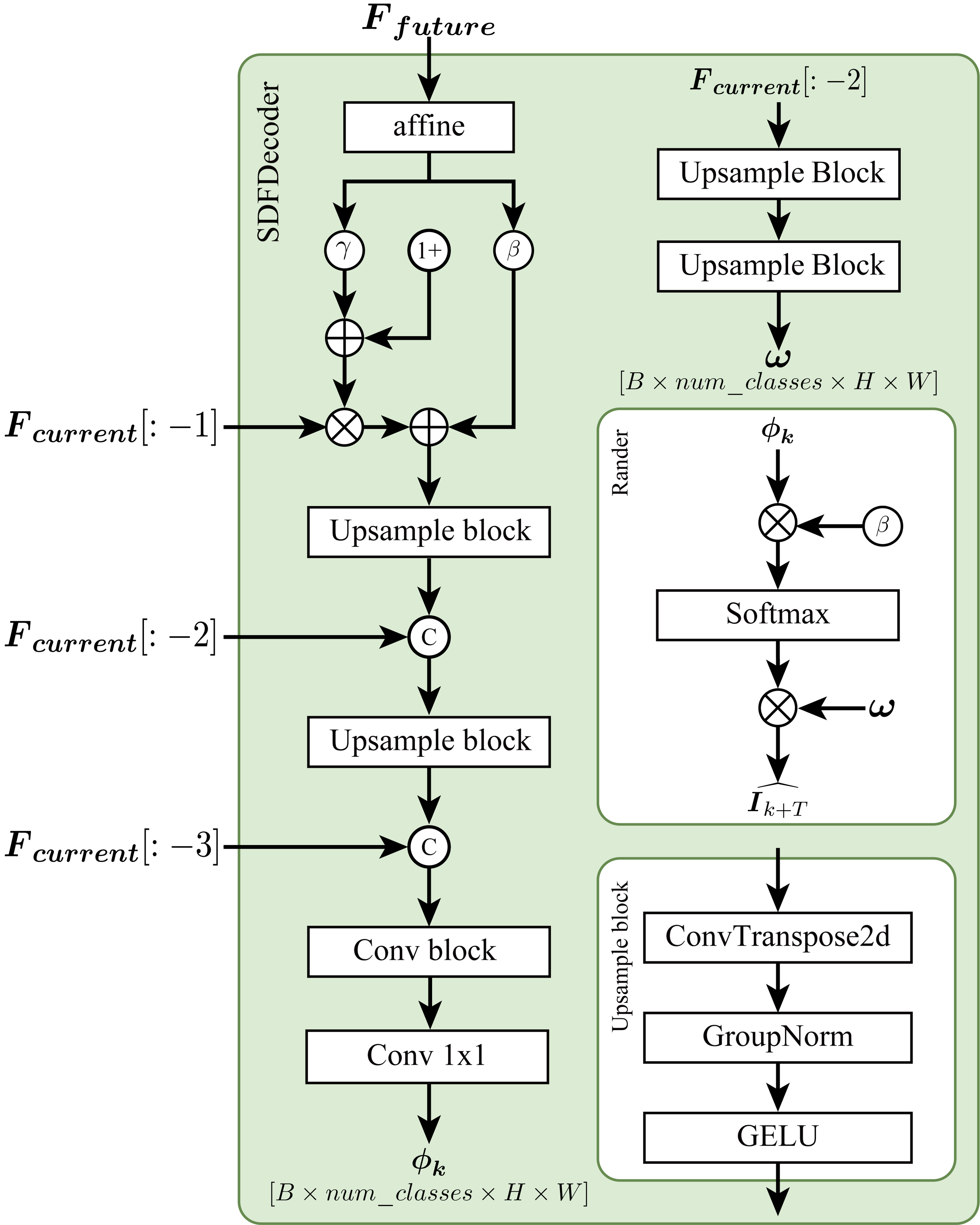}
\caption{Architecture of the auxiliary SDF decoder. The decoder combines temporally enriched visual features with the predicted future latent, refines them through skip-connected upsampling, and produces multi-class signed distance functions used for geometric supervision and differentiable rendering during training.}
\label{fig:sdf_decoder}
\end{figure}

\paragraph{SDF and rendering branch.}
The SDF decoder reconstructs a multi-class level-set field from the temporally enriched last-frame features and the future latent:
\begin{equation}
    \phisdf =
    \mathrm{Conv}_{1\times1}\left(
    \mathrm{Fuse}_2\left(
    \mathrm{Up}_2\left(
    \mathrm{Fuse}_1\left(
    \mathrm{Up}_1(\mathrm{AdaIN}(\mathbf{f}_2,\mathbf{z}_{t+K}))
    \oplus \mathbf{f}_1\right)\right)
    \oplus \mathbf{f}_0\right)\right).
\end{equation}
The detailed structure of the auxiliary SDF decoder is shown in \Cref{fig:sdf_decoder}.
It fuses the temporally enriched feature map with the predicted future latent and progressively upsamples the representation with skip connections, so that the SDF supervision preserves both molten pool geometry and fine keyhole/wire boundaries.

An appearance decoder predicts per-class textures, and a differentiable renderer converts the SDF into soft class probabilities:
\begin{equation}
    \alpha_c = \softmax(-\beta \phisdf)_c,\qquad
    \Ihat=\sum_{c=1}^{C}\alpha_c\odot\tau_c .
\end{equation}
The rendered image is compared with the observed future frame, providing photometric, gradient, Eikonal, mass, and texture-separation training signals.

\Cref{alg:lookahead_segmentation} summarizes the training and inference procedure of the proposed network.
The pseudocode follows the actual welding sensing pipeline: each mini-batch contains historical coaxial images, process parameters, aligned wire-state electrical signals, and future supervision targets; the deployment path uses only the encoder, process-parameter and wire-state electrical branches, temporal SSM, future predictors, motion cue, and mask decoder.

\begin{table}[t]
\centering
\refstepcounter{algorithm}
\label{alg:lookahead_segmentation}
\caption*{\textbf{Algorithm \thealgorithm:} Training -- Physics-Guided Lookahead Weld-Pool Segmentation}
\small
\setlength{\tabcolsep}{3pt}
\renewcommand{\arraystretch}{1.12}
\newcommand{\cmt}[1]{\hfill{\color{gray}\texttt{// #1}}}
\begin{tabular}{p{0.055\linewidth}p{0.895\linewidth}}
\toprule
\textbf{In}  & Training set $\mathcal{D}_{\mathrm{train}}$, validation set $\mathcal{D}_{\mathrm{val}}$, frame offsets $\mathcal{O}{=}\{-30,-20,-12,-6,-2,0\}$, prediction horizon $K$, augmentation set $\mathcal{A}$, model parameters $\theta$, loss weights $\Lambda$, learning rate $\alpha$, and epochs $E$. \\
\textbf{Out} & Best validation model $\theta^{\star}$ and future mask prediction $\mathbf{y}_{t+K}\in\R^{C\times H\times W}$ for keyhole, wire, and molten pool. \\
\midrule
0 & $\theta\leftarrow\mathrm{Init}(\,)$;\enspace load pretrained MiT-B1 \& frozen DINOv3-L/16 teacher \cmt{initialize} \\
1 & \textbf{for} $e=1,\ldots,E$ \textbf{do} \\
2 & \quad $\mathcal{B}\sim\mathcal{D}_{\mathrm{train}}$ \cmt{sample mini-batch} \\
3 & \quad \textbf{for} $\bigl(\{I_{t+o}\}_{o\in\mathcal{O}},\,\mathbf{p},\,\mathbf{c},\,M_{t+K},\,\boldsymbol{\phi}^{\mathrm{gt}}_{t+K},\,I_{t+K}\bigr)\in\mathcal{B}$ \textbf{do} \\
4 & \quad\quad $a\sim\mathcal{A}$;\enspace apply $a$ jointly to $\{I_{t+o}\},M_{t+K},\boldsymbol{\phi}^{\mathrm{gt}},I_{t+K}$ \cmt{consistent augmentation} \\
5 & \quad\quad $\mathbf{e}_p = \mathrm{MLP}(\mathbf{p})\in\R^{d/2}$ \cmt{process-parameter branch} \\
6 & \quad\quad $\mathbf{e}_c = \mathrm{MeanPool}\!\left(\mathrm{Mamba2}\!\left(\mathrm{Conv1D}_{1\to16\to d/2}(\mathbf{c})\right)\right)\in\R^{d/2}$ \cmt{wire-state branch} \\
7 & \quad\quad $\wphy = [\mathbf{e}_p\,;\,\mathbf{e}_c]\in\R^{d}$ \cmt{conditioning vector} \\
8 & \quad\quad \textbf{for} $o\in\mathcal{O}$ \textbf{do} \\
9 & \quad\quad\quad $\{\mathbf{s}_j\}_{j=1}^{4}\leftarrow\mathrm{MiT\text{-}B1}(I_{t+o})$;\enspace $\mathbf{f}_0\leftarrow\mathrm{Stem}(I_{t+o})$;\enspace $\mathbf{f}_1\leftarrow\mathrm{Down}_1(\mathbf{f}_0)$ \cmt{backbone + high-res stem} \\
10 & \quad\quad\quad $\mathbf{f}_2=\mathrm{Fuse}([P_j(\mathbf{s}_j)])$;\enspace $\mathbf{g}_{t+o}=\mathrm{MLP}_{\mathrm{fuse}}\!\left([\mathrm{Up}(Q_j\mathbf{f}_j)]\right)+\mathbf{f}_2$ \cmt{CondNorm + multi-scale fusion} \\
11 & \quad\quad \textbf{end for} \\
12 & \quad\quad $\bar{\mathbf{g}}_{t+o_i}=\mathbf{g}_{t+o_i}+\textstyle\sum_{c}\mathrm{Down}(\mathbf{m}_{t+o_{i-1},c})\,\mathbf{a}_c$ \cmt{prev-mask marker injection} \\
13 & \quad\quad $\mathbf{X}^{(\ell+1)}=\mathrm{Mamba2}\!\left(\mathrm{WinPart}\!\left(\mathrm{Shift}^{[\ell\,\mathrm{odd}]}_{s}(\mathbf{X}^{(\ell)})\right)\right),\;\ell=0,1,2,3$ \cmt{4 shifted-window layers} \\
   & \quad\quad $\mathbf{z}_t=P_z(\mathrm{AvgPool}(\tilde{\mathbf{g}}_t))$;\enspace $\mathbf{Z}_{\mathrm{hist}}=\{\mathrm{Pool}_{4\times8}(\bar{\mathbf{g}}_{t+o_i})\}_{i=1}^{T}$ \cmt{current latent + token grid} \\
14 & \quad\quad $\mathbf{q}^{(0)}_K=\mathbf{z}_t+\mathbf{e}_K$;\enspace $\mathbf{q}^{(r+1)}_K=\mathrm{AttnBlock}(\mathbf{q}^{(r)}_K,\,\mathbf{Z}_{\mathrm{hist}})$;\enspace $\mathbf{z}_{t+K}=\mathbf{z}_t+\Delta(\mathbf{q}_K)$ \cmt{future latent} \\
15 & \quad\quad $\mathbf{f}^{\mathrm{future}}_2=\tilde{\mathbf{g}}_t+\rho\cdot\Delta_{\mathrm{dense}}\!\left([\tilde{\mathbf{g}}_t,\,\mathbf{g}_t,\,\overline{\mathbf{g}},\,\mathbf{g}_t-\mathbf{g}_{t-T+1},\,\mathbf{u}_K]\right)$ \cmt{dense future feature} \\
16 & \quad\quad $(\mathbf{c}_K,\,R_K,\,\boldsymbol{\omega}_K,\,\hat{\mathbf{c}}_K)\leftarrow\mathrm{KeyholeMotionHead}(\mathbf{z}_t,\mathbf{Z}_{\mathrm{hist}},\tilde{\mathbf{g}}_t,\mathbf{e}_K)$ \cmt{circle center/radius/ang.\ velocity + future center; $(\mathbf{c}_K,R_K,\boldsymbol{\omega}_K)$ supervised by $\mathcal{L}_{\mathrm{motion}}$} \\
   & \quad\quad $\mathbf{f}^{\mathrm{future}}_2\leftarrow\mathrm{MotionXAttn}\!\left(\mathbf{f}^{\mathrm{future}}_2,\,\mathcal{G}(\hat{\mathbf{c}}_K,R_K)\right)$ \cmt{bias dense feature toward predicted keyhole via Gaussian prior} \\
17 & \quad\quad $\mathbf{r}_K=\mathrm{XAttn}\!\left(\mathrm{Conv}\!\left((1{+}\gamma(\mathbf{z}_{t+K}))\odot\mathbf{f}^{\mathrm{future}}_2+\beta(\mathbf{z}_{t+K})\right),\,\mathbf{z}_{t+K}\right)$ \\
   & \quad\quad $\mathbf{y}_{t+K}=\mathrm{MaskHead}(\mathbf{r}_K)\in\R^{C\times H\times W}$ \cmt{future segmentation mask} \\
18 & \quad\quad $\boldsymbol{\phi}_{\mathrm{sdf}}=\mathrm{SDFDecoder}(\tilde{\mathbf{g}}_t,\mathbf{z}_{t+K})$;\enspace $\hat{I}_{t+K}={\textstyle\sum_c}\,\mathrm{softmax}({-}\beta\boldsymbol{\phi}_{\mathrm{sdf}})_c\odot\tau_c$ \cmt{SDF field + differentiable render} \\
19 & \quad\quad $\mathcal{L}=\textstyle\sum_i\lambda_i\mathcal{L}_i\quad\text{(Eq.~\ref{eq:total_loss})}$ \cmt{seg / SDF / render / align / vid / distill / state / f1 / hbs / kh / motion / dense} \\
20 & \quad \textbf{end for} \\
21 & \quad $\theta\leftarrow\theta-\alpha\nabla_\theta\mathcal{L}$ \cmt{AdamW step} \\
22 & \quad \textbf{if} $\mathrm{mIoU}(\mathcal{D}_{\mathrm{val}})$ improves \textbf{then}\enspace $\theta^{\star}\leftarrow\theta$ \cmt{save best checkpoint} \\
23 & \textbf{end for} \\
24 & \textbf{return} $\theta^{\star}$;\enspace inference: $\mathbf{y}_{t+K}\leftarrow\mathrm{Forward}(\{I_{t+o}\},\mathbf{p},\mathbf{c},K)$ \cmt{deploy without SDF/render/distill heads} \\
\bottomrule
\end{tabular}
\end{table}

\subsection{Training Objective}
\label{sec:training}

The training objective is designed to supervise future segmentation, molten pool geometry, temporal consistency, and cross-condition feature stability without requiring the auxiliary branches at deployment.
These terms correspond to welding-specific requirements: accurate region masks for online perception, smooth SDF geometry for pool boundaries, temporal consistency for continuous process evolution, and feature stability across different process-parameter settings.
The objective combines pixel-level mask/SDF supervision on annotated frames, differentiable SDF rendering on temporally adjacent frames, MMD-based feature alignment across welding conditions, DINOv3 feature distillation, and future-specific auxiliary losses.
The overall loss is written as
\begin{align}
    \cL
    ={}& \lambda_{\mathrm{seg}}\cL_{\mathrm{seg}}
    + \lambda_{\mathrm{sdf}}\cL_{\mathrm{sdf}}
    + \lambda_{\mathrm{render}}\cL_{\mathrm{render}}
    + \lambda_{\mathrm{align}}\cL_{\mathrm{align}}
    + \lambda_{\mathrm{vid}}\cL_{\mathrm{vid}}
    + \lambda_{\mathrm{distill}}\cL_{\mathrm{distill}} \nonumber\\
    &+ \lambda_{\mathrm{state}}\cL_{\mathrm{state}}
    + \lambda_{\mathrm{f1}}\cL_{\mathrm{f1}}
    + \lambda_{\mathrm{hbs}}\cL_{\mathrm{hbs}}
    + \lambda_{\mathrm{kh}}\cL_{\mathrm{kh}}
    + \lambda_{\mathrm{motion}}\cL_{\mathrm{motion}}
    + \lambda_{\mathrm{dense}}\cL_{\mathrm{future\_dense}} .
    \label{eq:total_loss}
\end{align}
The supervised mask/SDF terms use focal and boundary losses with online hard-example mining.
The rendering term compares the SDF-rendered image with the observed future frame using photometric, gradient, Eikonal, mass, and texture-separation constraints.
The alignment term uses MMD on latent features to reduce feature drift across welding parameter settings, while the video consistency term encourages temporally adjacent predictions to remain compatible with the observed image sequence.
A frozen DINOv3 ViT-L/16 teacher~\cite{dinov3_2025} supplies dense feature distillation targets through the projection head attached to the image encoder.
Fine-scale auxiliary losses supervise the stride-2 feature map, a region-smoothing branch, and a cropped keyhole refinement head.
The keyhole motion loss uses valid samples whose fitted circular motion satisfies minimum-radius and residual thresholds.
All teacher, projection, rendering, and auxiliary heads are removed or ignored at deployment; inference uses the image encoder, process-parameter and wire-state electrical branches, temporal SSM, future predictors, and mask decoder.

\begin{table*}[t]
\centering
\caption{Loss weights and key thresholds used in the final configuration.}
\label{tab:loss_hyperparameters}
\vspace{4pt}
\scriptsize
\setlength{\tabcolsep}{4pt}
\begin{tabular*}{\textwidth}{@{\extracolsep{\fill}}p{0.25\textwidth}p{0.39\textwidth}p{0.28\textwidth}@{}}
\toprule
Group & Terms & Values \\
\midrule
Primary mask supervision & focal, Dice, boundary & weights $(1.0,1.0,0.2)$; OHEM ratio $0.75$ \\
Auxiliary SDF supervision & focal, Dice, boundary & weights $(0.2,0.5,0.3)$; OHEM ratio $1.0$ \\
SDF rendering & photometric, gradient, Eikonal, mass, separation & weights $(1.0,0.5,0.2,0.2,0.05)$; renderer $\beta$ initialized to $15.0$ and clamped to $[1,80]$ \\
Auxiliary regularizers & domain alignment, pseudo-label, target supervision, DINOv3 distillation, state loss & $\lambda_{da}=0.1$, $\lambda_{pl}=0.2$, $\lambda_{tgt}=0.5$, $\lambda_{distill}=0.1$, $\lambda_{state}=0.1$ \\
Dense and keyhole auxiliaries & fine-scale mask, history mask, smoothing branch, keyhole mask, keyhole center, keyhole motion & $(0.3,0.1,0.3,0.5,0.2,0.05)$ \\
Keyhole-motion validity & minimum valid centers, radius range, residual threshold, maximum angular velocity & $4$, $[0.005,0.75]$, $0.04$, $\pi$ \\
\bottomrule
\end{tabular*}
\end{table*}

\section{Experimental Verification and Analysis}
\label{sec:experiments}

\subsection{Experimental Settings}

The experimental verification is conducted on the 43-sequence laser welding dataset described in \Cref{sec:data}.
All models are trained and evaluated using the same sequence-level split, so the validation sequences correspond to welding conditions that are not seen as identical frame sequences during training.
\revone{The 9 validation sequences are used as a controlled process-level validation split for comparing model variants and welding-relevant design choices.}
\revone{This protocol is intended to evaluate whether the proposed visual--process--sensor representation generalizes across unseen welding sequences within the collected process window; external cross-platform testing is treated as a deployment-oriented extension rather than the main scope of this study.}
The input image size is $480\times480$ in the final configuration.
The model uses six historical frames with offsets $[-30,-20,-12,-6,-2,0]$ and predicts the frame $K$ steps after the anchor frame.
This setting evaluates future molten pool segmentation rather than current-frame mask fitting, and the effect of $K$ is analyzed separately in \Cref{sec:horizon_analysis}.

The experimental framework in this study is implemented in PyTorch.
Training is conducted on a workstation equipped with an Intel(R) Xeon(R) Platinum 8358 CPU, three NVIDIA A800 GPUs, and 256\,GB of memory.
We use the AdamW optimizer with an initial learning rate of $10^{-4}$.
The training schedule contains 64 epochs, matching the final supervised training schedule used for the model comparison experiments.
The batch size is set to 63 over the available GPUs, and \revone{the checkpoint with the best validation performance is used for the reported validation comparisons}.
The training script uses bfloat16 mixed-precision computation, while the loss terms are computed with the same mask and SDF targets as the final model.
Unless otherwise stated, all IoU and mIoU values in this section are reported as percentages.
\revone{The experiments use a fixed split seed and a single training run for each reported configuration; consequently, very small differences are discussed as trends only when they are consistent with the welding-process interpretation.}

\subsection{Progressive Component Ablation}
\label{sec:component_ablation}

For compact notation in \Cref{tab:component_progressive_ablation}, ``Proc. cond.'' and ``Sensor cond.'' denote process-parameter and wire-state electrical conditioning, respectively; MS, SSM, SDF, and KD denote multi-scale fusion, state space modeling, signed distance function supervision, and knowledge distillation, respectively. All metrics in this table are reported as percentages, with the best and runner-up entries marked in bold and underlined.

\begin{table*}[t]
\centering
\caption{Progressive component ablation for the proposed lookahead segmentation network. Class-wise columns report IoU for each welding target region, and mIoU is the mean over keyhole, wire, and molten pool.}
\label{tab:component_progressive_ablation}
\vspace{4pt}
\scriptsize
\setlength{\tabcolsep}{4pt}
\begin{tabular*}{\textwidth}{@{\extracolsep{\fill}}c l c c c c@{}}
\toprule
No. & Method & mIoU (\%) $\uparrow$ & IoU$_{kh}$ (\%) $\uparrow$ & IoU$_{wire}$ (\%) $\uparrow$ & IoU$_{pool}$ (\%) $\uparrow$ \\
\midrule
1  & Baseline & 49.20 & 20.94 & 63.25 & 63.41 \\
2  & + Proc. cond. & 50.78 & 21.12 & 64.28 & 66.93 \\
3  & + Sensor cond. & 51.68 & 22.04 & 65.72 & 67.27 \\
4  & + MS fusion & 53.10 & 24.26 & 66.50 & 68.54 \\
5  & + Hist. frames & 54.35 & 24.77 & 67.51 & 70.76 \\
6  & + Patch SSM & 62.43 & 30.15 & 77.55 & 79.59 \\
7  & + Future decoder & 67.47 & 37.01 & 81.78 & 83.62 \\
8  & + Dense future & 68.64 & 38.54 & 82.51 & 84.86 \\
9  & + Horizon latent & 71.06 & 40.77 & 85.30 & 87.12 \\
10 & + Keyhole motion & 72.51 & 45.06 & 85.32 & 87.15 \\
11 & + SDF & 73.30 & 45.71 & 86.07 & 88.13 \\
12 & + KD & \underline{74.46} & \underline{46.80} & \underline{87.23} & \underline{89.35} \\
13 & WeldMamba (+ GELU) & \textbf{74.63} & \textbf{46.86} & \textbf{87.61} & \textbf{89.42} \\
\bottomrule
\end{tabular*}
\end{table*}

\begin{figure}[H]
\centering
\includegraphics[width=0.68\textwidth]{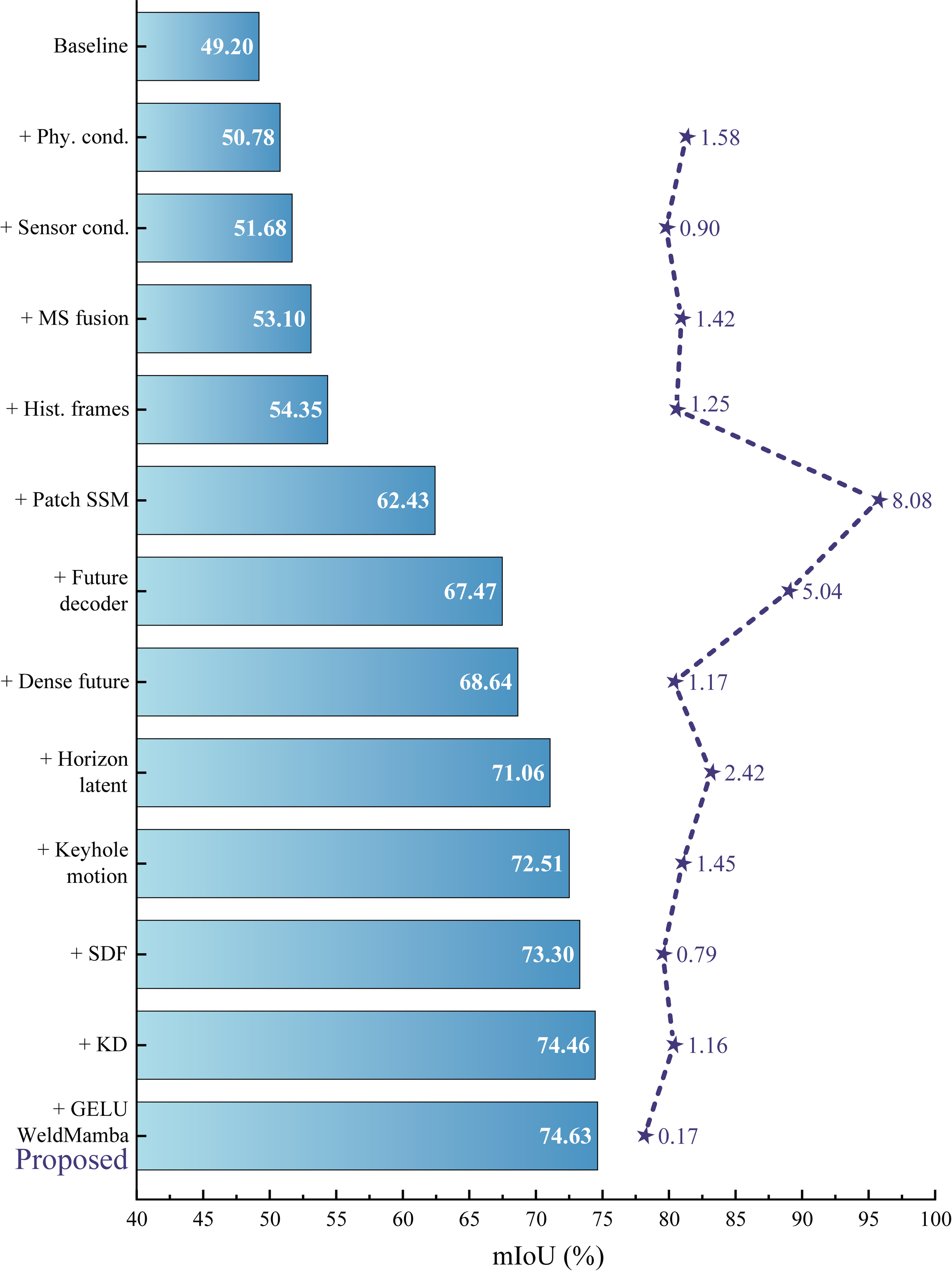}
\caption{Contribution of each component to mIoU in the ablation experiment.}
\label{fig:component_ablation_visual}
\end{figure}

\Cref{fig:component_ablation_visual} shows the contribution of each component to mIoU in the ablation experiment.

\begin{figure}[H]
\centering
\includegraphics[width=0.98\textwidth]{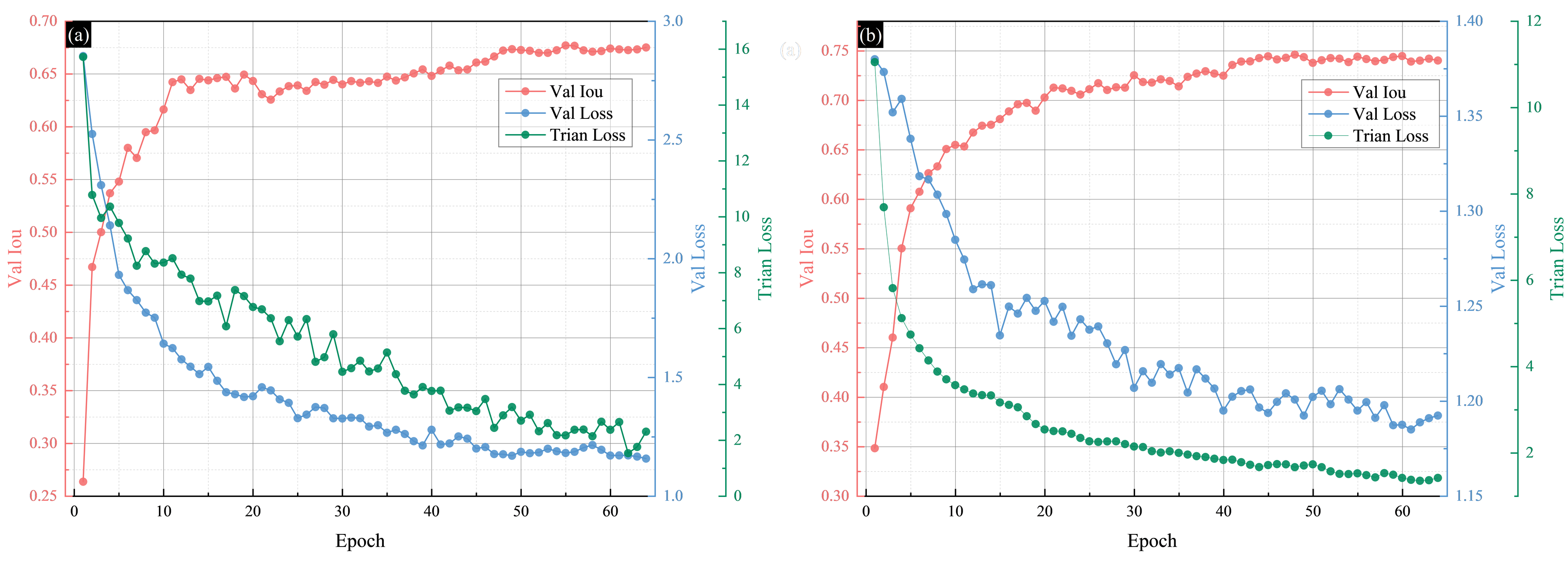}
\caption{Training curves for representative configurations in the progressive component ablation. Panels (a) and (b) correspond to No.~7 and No.~13 in \Cref{tab:component_progressive_ablation}, respectively.}
\label{fig:component_training_curves}
\end{figure}

\Cref{fig:component_training_curves} shows the training curves of No.~7 and No.~13 as representative intermediate and full configurations in the progressive ablation.

\Cref{tab:component_progressive_ablation} should be read as an error-diagnosis table rather than as a list of architectural parts.
No.~1 is an image-only prediction model built from a grayscale MiT-B1 encoder and a plain mask head, so it can only infer information from instantaneous appearance.
This baseline already obtains relatively good IoU values for the wire and molten pool, but its keyhole IoU is only 20.94\%.
This imbalance is consistent with coaxial laser wire-feed images: the wire and molten pool occupy larger continuous regions, whereas the keyhole is a small opening whose visible contour can be shifted by recoil pressure, plume disturbance, specular reflection, and beam oscillation.
Therefore, the first bottleneck is not whether the network can recognize the three semantic categories, but whether it can localize the smallest and most unstable physical region before it moves or becomes partially occluded.

No.~2 and No.~3 introduce process-parameter conditioning and wire-state electrical-signal conditioning, increasing mIoU from 49.20\% to 51.68\%.
This indicates that operating-condition information reduces cross-condition ambiguity: similar gray-level patterns can be produced under different spot/ring power ratios, wire-feeding speeds, and oscillation settings, while the same semantic region may appear differently when heat input or wire transfer changes.
This has also been observed in welding studies that combine visual information with process parameters or physical constraints \citep{li2026multitask_spatiotemporal,li2026simphysnet}, but such information alone cannot resolve the rapid displacement of the keyhole.
No.~4 adds multi-scale visual fusion and raises mIoU to 53.10\%, showing that richer image morphology helps retain small keyhole and wire structures while still using wider molten pool context.
No.~5 further adds historical frames and reaches 54.35\%, indicating that recent visual evolution begins to provide motion cues, although the keyhole remains the dominant source of error.

From No.~5 to No.~6, mIoU increases by 8.08 percentage points, and the wire and molten pool IoU values reach 77.55\% and 79.59\%, respectively.
Such a large jump indicates that many earlier false predictions were caused by transient visual events rather than by category confusion: reflections, plume interference, and short-lived transfer states can resemble region boundaries in a single frame, while their persistence across time is different from that of the actual wire and molten pool.
This agrees with prior laser-welding monitoring work showing that time-series keyhole and molten pool signals are decisive for state recognition \citep{yu2022_dynamic_pool_cnn_lstm,zhao2024_continuous_video,yan2025_lsw_crossvit}.
The keyhole remains much lower than the other two classes, which is also expected because a few-pixel displacement or a momentary occlusion causes a large IoU penalty for such a small target.

No.~7--13 show that once the main temporal ambiguity is reduced, the remaining gains become localized.
The increase from No.~6 to No.~9 mainly improves the delayed-frame layout, indicating that the future wire, keyhole, and pool positions cannot be treated as a rigid shift of the current mask.
The jump from No.~9 to No.~10 is concentrated on the keyhole, from 40.77\% to 45.06\%, while the wire and pool barely change; this class-selective improvement supports the view that the residual error is dominated by keyhole motion rather than by global mask quality.
No.~11--13 add SDF supervision, feature distillation, and the final GELU activation change after the main temporal and future-prediction components are already present. \revone{These late-stage increments are comparatively small and should be interpreted as refinement trends under the single-run protocol, not as independently isolated effects.}
Thus, the cumulative ablation supports a welding-specific conclusion: lookahead segmentation is limited first by temporal instability of the process, especially around the keyhole, and only secondarily by static appearance ambiguity or boundary smoothness.

\subsection{Comparison with Alternative Network Replacements}
\label{sec:network_replacement_comparison}

To compare the performance gap between the proposed spatiotemporal module and existing models, we conduct a controlled module-replacement experiment with SegFormer~\citep{xie2021segformer}, CFFM~\citep{sun2022cffm}, MRCFA~\citep{sun2022mrcfa}, Mask2Former~\citep{cheng2022mask2former}, and DeepLabv3+~\citep{chen2018deeplabv3plus}.
In each variant, only the PatchTemporalSSM module is replaced by the compared network, while the history offsets, process and sensor conditioning, future prediction modules, decoder, loss functions, and sequence-level train--validation split remain unchanged.
The comparison therefore evaluates whether these alternatives provide a better temporal representation inside the same lookahead welding framework, rather than comparing standalone current-frame segmentation models.
\revone{Consequently, lower values for architectures such as Mask2Former and DeepLabv3+ should not be interpreted as their best-case standalone performance.}

\begin{table}[t]
\centering
\caption{Model accuracies of different network replacements. Best results are displayed in bold and runner-up results are underlined.}
\label{tab:network_replacement_comparison}
\vspace{4pt}
\scriptsize
\setlength{\tabcolsep}{5pt}
\begin{tabular}{lcccc}
\toprule
Method & mIoU (\%) $\uparrow$ & IoU$_{kh}$ (\%) $\uparrow$ & IoU$_{wire}$ (\%) $\uparrow$ & IoU$_{pool}$ (\%) $\uparrow$ \\
\midrule
SegFormer & 71.53 & 42.15 & 85.07 & 87.37 \\
CFFM & 72.46 & 44.73 & 85.25 & 87.41 \\
MRCFA & \underline{73.51} & \underline{45.59} & \underline{86.91} & \underline{88.02} \\
Mask2Former & 61.29 & 39.18 & 72.61 & 72.09 \\
DeepLabv3+ & 60.50 & 38.92 & 71.23 & 71.36 \\
WeldMamba (ours) & \textbf{74.63} & \textbf{46.86} & \textbf{87.61} & \textbf{89.42} \\
\bottomrule
\end{tabular}
\end{table}

\Cref{tab:network_replacement_comparison} shows that the proposed PatchTemporalSSM is not merely benefiting from a stronger generic segmentation backbone.
CFFM and MRCFA perform better than single-frame segmentation replacements because their feature-fusion designs can use cross-frame cues, but WeldMamba still gives the highest mIoU and the best class-wise IoU values.
The advantage is most visible on the keyhole, where a small spatial displacement produces a large IoU penalty and the local welding dynamics are easily confused with reflection or plume-induced intensity changes.
Mask2Former and DeepLabv3+ are weaker in this controlled setting because their strengths in mask decoding and atrous context aggregation do not directly model the non-uniform historical evolution required for future molten pool segmentation.

\subsection{Temporal History Ablation}
\label{sec:temporal_history_ablation}

\begin{table}[t]
\centering
\caption{History-length ablation with contiguous historical frames. Offsets are listed from newest to oldest relative to the anchor frame, and temporal span is reported in milliseconds at 50\,fps. Best results are displayed in bold and runner-up results are underlined.}
\label{tab:history_length_ablation}
\vspace{4pt}
\scriptsize
\setlength{\tabcolsep}{4pt}
\begin{tabular}{p{0.29\linewidth}ccccc}
\toprule
History offsets & Span [ms] & mIoU (\%) & IoU$_{kh}$ (\%) & IoU$_{wire}$ (\%) & IoU$_{pool}$ (\%) \\
\midrule
$\{0\}$ & 0 & 25.85 & 7.21 & 34.42 & 35.92 \\
$\{0,-1\}$ & 20 & 28.06 & 8.32 & 36.07 & 39.78 \\
$\{0,-1,-2\}$ & 40 & 37.01 & 13.11 & 49.84 & 48.08 \\
$\{0,-1,-2,-3\}$ & 60 & 45.11 & 19.92 & 57.45 & 57.97 \\
$\{0,-1,-2,-3,-4\}$ & 80 & 51.29 & 22.06 & 68.73 & 63.09 \\
$\{0,-1,-2,-3,-4,-5\}$ & 100 & \underline{56.89} & \underline{28.82} & \underline{71.53} & \underline{70.33} \\
$\{0,-1,-2,-3,-4,-5,-6\}$ & 120 & \textbf{58.40} & \textbf{30.96} & \textbf{72.59} & \textbf{71.64} \\
\bottomrule
\end{tabular}
\end{table}

\Cref{tab:history_length_ablation} shows why an instantaneous image is insufficient for a predictive welding monitor.
With only the anchor frame, mIoU is 25.85\% and keyhole IoU remains low at 7.21\%, because the image gives position but not velocity or phase.
For example, a small bright-dark opening may correspond to a keyhole that is shrinking, reopening, or being swept by oscillating heat input; these cases can look similar at one instant but lead to different future masks.
Adding contiguous frames supplies this missing short-term evolution, and all three classes improve monotonically up to the 120 ms span.
The strongest relative change appears on the keyhole, while the wire and molten pool also benefit because recent frames help separate real wire transfer from bright pool reflections and prevent the pool boundary from being fitted to temporary surface fluctuations.
The small gain from 100 ms to 120 ms shows that dense nearby frames are useful but begin to saturate; after the recent motion has been observed, the remaining uncertainty is more related to slower thermal history than to another adjacent frame.

\begin{table}[t]
\centering
\caption{Frame-spacing ablation with six input frames. Offsets are listed from newest to oldest relative to the anchor frame, and temporal span is reported in milliseconds at 50\,fps. Best results are displayed in bold and runner-up results are underlined.}
\label{tab:history_spacing_ablation}
\vspace{4pt}
\scriptsize
\setlength{\tabcolsep}{4pt}
\begin{tabular}{p{0.29\linewidth}ccccc}
\toprule
History offsets & Span [ms] & mIoU (\%) & IoU$_{kh}$ (\%) & IoU$_{wire}$ (\%) & IoU$_{pool}$ (\%) \\
\midrule
$\{0,-1,-2,-3,-4,-5\}$ & 100 & 56.89 & 28.82 & 71.53 & 70.33 \\
$\{0,-2,-4,-6,-8,-10\}$ & 200 & 60.05 & 31.59 & 75.41 & 73.15 \\
$\{0,-3,-6,-9,-12,-15\}$ & 300 & 65.05 & 35.99 & 79.13 & 80.03 \\
$\{0,-4,-8,-12,-16,-20\}$ & 400 & \underline{69.50} & \underline{42.59} & \underline{82.23} & \underline{83.69} \\
$\{0,-2,-6,-12,-20,-30\}$ & 600 & \textbf{74.63} & \textbf{46.86} & \textbf{87.61} & \textbf{89.42} \\
\bottomrule
\end{tabular}
\end{table}

\Cref{tab:history_spacing_ablation} separates the effect of temporal span from the effect of simply increasing the number of frames.
When the number of frames is fixed at six, extending the span from 100 ms to 400 ms raises mIoU from 56.89\% to 69.50\%, and the non-uniform 600 ms setting reaches 74.63\%.
The molten pool shows the clearest gain, increasing from 70.33\% to 89.42\%, because molten pool shape has a longer thermal memory than the keyhole opening.
A narrow contiguous window can observe local motion, but it may fail to determine whether the molten pool is gradually expanding, contracting, or being biased by heat accumulated over previous oscillation cycles.
The keyhole also improves from 28.82\% to 46.86\%, indicating that a longer observation range does not discard useful recent dynamics when the sampling still includes frames near the anchor.
Taken together, these two ablation experiments lead to a clear conclusion: the history-length study evaluates whether the network benefits from observing more contiguous frames before the anchor frame, whereas the frame-spacing study fixes the number of input frames and changes how far the historical information extends into the past.
This distinction is important for laser wire-feed welding because keyhole behavior and droplet-transfer dynamics require dense recent frames, whereas molten pool deformation and heat accumulation benefit from a longer temporal span.

\subsection{Prediction Horizon Analysis}
\label{sec:horizon_analysis}

The prediction horizon $K$ determines the time margin available to a downstream controller.
Short horizons mostly evaluate whether the current process state can be tracked, whereas long horizons test whether the visual and sensor history contains enough information to anticipate the next welding state before actuation delay becomes relevant.
To make this accuracy--lookahead trade-off explicit, the horizon analysis uses several values of $K$ with the same sequence-level validation split and the same six-frame history.
\revone{The 500 ms setting is the deployment-oriented target because the final inference path is exported to ONNX and measured on an NVIDIA RTX 2080 Ti with an average latency of 485.63 ms per sample under the adopted input setting. The 500 ms row therefore approximately compensates for the measured computation delay of the deployed model, whereas shorter horizons are mechanically easier near-frame forecasts and should not be read as the most useful control setting.}

\begin{table}[t]
\centering
\caption{Prediction-horizon accuracy analysis. Time is reported in milliseconds at 50\,fps. Best results are displayed in bold and runner-up results are underlined.}
\label{tab:horizon_accuracy}
\vspace{4pt}
\scriptsize
\setlength{\tabcolsep}{4pt}
\begin{tabular}{crrrrr}
\toprule
$K$ & Time [ms] & mIoU (\%) & IoU$_{kh}$ (\%) & IoU$_{wire}$ (\%) & IoU$_{pool}$ (\%) \\
\midrule
1  & 20   & \textbf{90.56} & \textbf{75.87} & \textbf{98.05} & \textbf{97.76} \\
5  & 100  & \underline{86.66} & \underline{69.89} & \underline{94.74} & \underline{95.36} \\
10 & 200  & 78.87 & 54.11 & 90.90 & 91.60 \\
25 & 500  & 74.63 & 46.86 & 87.61 & 89.42 \\
50 & 1000 & 65.19 & 40.15 & 75.63 & 79.78 \\
\bottomrule
\end{tabular}
\end{table}

\Cref{tab:horizon_accuracy} quantifies the price paid for earlier control information.
At 20 ms and 100 ms, the predicted masks are still close to the observed welding state, and mIoU remains above 86\%.
\revone{At 500 ms, which is close to the measured 485.63 ms ONNX inference latency on an NVIDIA RTX 2080 Ti, the full model still reaches 74.63\% mIoU, indicating that the learned history remains informative over the delay scale relevant to the exported inference pipeline.}
At 1000 ms, however, mIoU decreases to 65.19\%, suggesting that stochastic transfer behavior, plume disturbance, and accumulated thermal variation begin to dominate over deterministic visual extrapolation.
The class-wise trend is especially important: keyhole IoU decreases from 75.87\% to 40.15\% across the table, while the wire and molten pool remain above 75\% even at 1000 ms.
This means that long-horizon uncertainty is concentrated at the most localized laser-material interaction region; the larger regions retain more predictable shape continuity and are therefore less sensitive to the selected horizon.

\subsection{Temporal Depth Ablation}
\label{sec:patch_temporal_depth_ablation}

This experiment tests whether the remaining errors are mainly due to insufficient temporal capacity or to the intrinsic variability of the welding process.
The number of PatchTemporalSSM blocks is varied while keeping the same sequence-level split, history offsets, prediction horizon, image size, optimizer, and loss configuration.

\begin{table}[t]
\centering
\caption{PatchTemporalSSM depth ablation. Best results are displayed in bold and runner-up results are underlined.}
\label{tab:patch_temporal_blocks}
\vspace{4pt}
\scriptsize
\setlength{\tabcolsep}{5pt}
\begin{tabular}{ccccc}
\toprule
PatchTemporalSSM blocks & mIoU (\%) $\uparrow$ & IoU$_{kh}$ (\%) $\uparrow$ & IoU$_{wire}$ (\%) $\uparrow$ & IoU$_{pool}$ (\%) $\uparrow$ \\
\midrule
1 & 54.88 & 32.11 & 71.86 & 72.66 \\
2 & 61.99 & 36.87 & 75.74 & 76.36 \\
3 & 67.94 & 41.48 & 79.92 & 82.42 \\
4 & \underline{74.63} & \underline{46.86} & \underline{87.61} & \underline{89.42} \\
5 & \textbf{74.92} & \textbf{46.95} & \textbf{87.86} & \textbf{89.94} \\
\bottomrule
\end{tabular}
\end{table}

\Cref{tab:patch_temporal_blocks} shows a clear capacity-saturation trend.
From one to four blocks, mIoU increases from 54.88\% to 74.63\%, and all three target regions improve steadily.
This suggests that shallow temporal reasoning can already capture coarse wire and molten pool morphology, but additional PatchTemporalSSM blocks are needed to align the keyhole, wire tip, and pool boundary when they evolve with different characteristic speeds.
The fifth block gives the best numerical result, but the gain over four blocks is only 0.29 percentage points in mIoU, with a keyhole improvement of only 0.09 percentage points.
\revone{We do not claim this small difference to be statistically significant under the single-run protocol.}
Such a small change indicates that the dominant remaining errors are no longer solved by simply increasing temporal depth; they are more likely associated with ambiguous frames, abrupt transfer events, and annotation-level boundary uncertainty.
The four-block configuration is therefore used as the final setting because it reaches nearly the same accuracy while avoiding extra computation that brings little additional welding-state information.

\subsection{Visualization}
\label{sec:attention_visualization}

\begin{figure}[H]
\centering
\includegraphics[width=0.98\textwidth]{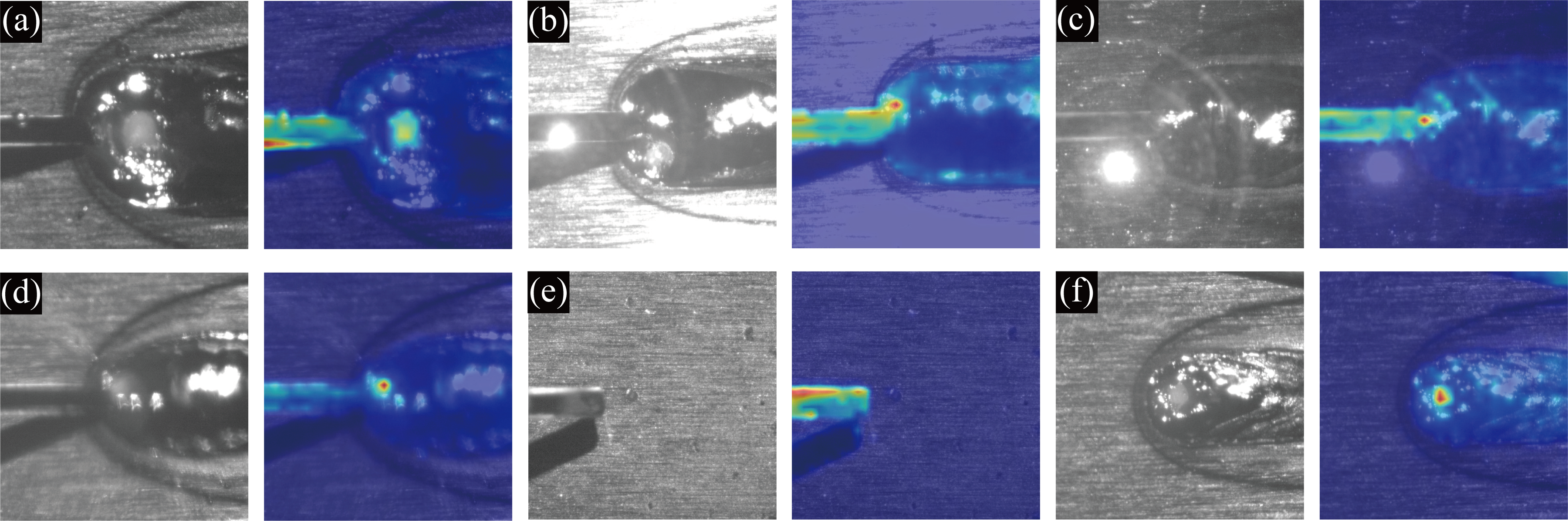}
\caption{Attention visualization under representative welding-image conditions produced by Grad-CAM: (a) normal illumination, (b) high illumination, (c) strong glare interference, (d) metal-vapor interference, (e) wire-only image, and (f) image containing only keyhole and molten pool regions.}
\label{fig:attention_visualization}
\end{figure}

To complement the quantitative results, we visualize part of the model outputs to show attention allocation and sequence-level prediction behavior under typical welding-image conditions.
These examples cover normal illumination, high illumination, glare, metal-vapor interference, class-limited scenes, and ordered future predictions from $T$ to $T+100$\,ms.
Taken together, they indicate that the model focuses on physically meaningful welding structures and maintains temporal coherence, while the remaining errors are concentrated on the smallest and most unstable regions.

\paragraph{Attention weight visualization.}
During training, Grad-CAM~\citep{selvaraju2017gradcam} is used to visualize the attention distribution of the visual backbone.
\Cref{fig:attention_visualization} shows that under normal illumination and high brightness, the response concentrates on the wire-entry region, the keyhole neighborhood, and the molten pool interior rather than on the surrounding plate texture.
Under strong glare and metal-vapor interference, the attention is still centered on the transition among the wire, keyhole, and molten pool instead of being dominated by the brightest saturated area.
In the class-limited scenes, the response remains on the visible welding structures and suppresses irrelevant background, which indicates that the learned representation tracks welding-relevant cues rather than incidental appearance.

\begin{figure}[H]
\centering
\includegraphics[width=0.98\textwidth]{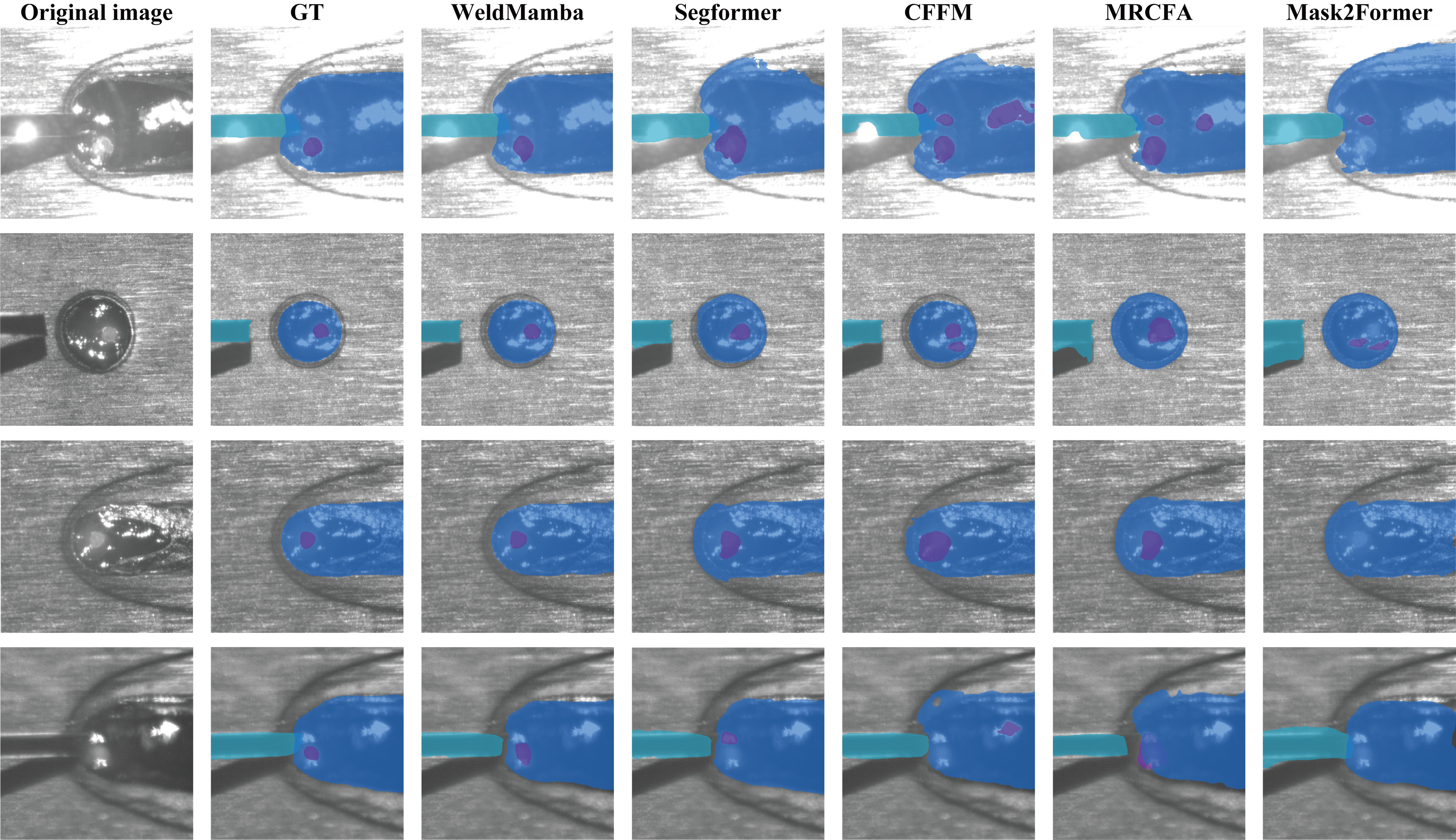}
\caption{Qualitative comparison of predicted masks under representative welding-image conditions. From left to right: original image, ground truth, WeldMamba, SegFormer, CFFM, MRCFA, and Mask2Former.}
\label{fig:model_inference_visualization}
\end{figure}

\paragraph{Different model inference visualization.}
\Cref{fig:model_inference_visualization} compares the final masks predicted by different networks.
Most methods recover the coarse molten pool when contrast is clear, but they diverge around the small keyhole and thin wire tip, where local displacement, partial occlusion, and transient highlights readily cause contour shifts.
WeldMamba preserves the coupling among the keyhole, wire, and molten pool more consistently than the other models, which reduces implausible shape breaks and better matches the underlying welding state.
This is important for lookahead segmentation because a plausible pool outline alone is insufficient if the keyhole or wire-entry geometry is misplaced.

\begin{figure}[H]
\centering
\includegraphics[width=0.98\textwidth]{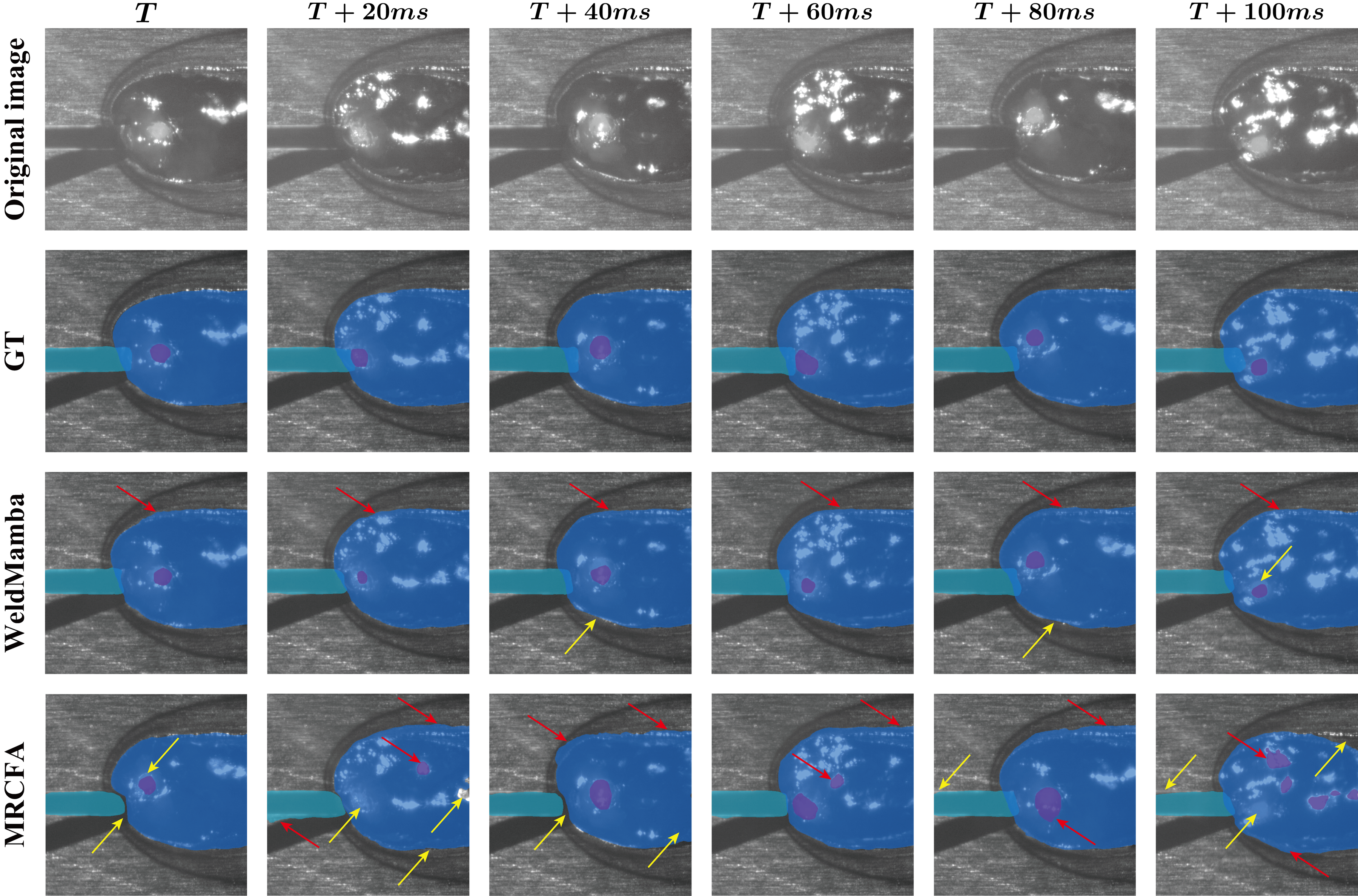}
\caption{Sequential inference visualization from $T$ to $T+100$\,ms. Red arrows indicate over-segmentation and yellow arrows indicate under-segmentation.}
\label{fig:sequential_inference_visualization}
\end{figure}

\paragraph{Sequential consistency.}
\Cref{fig:sequential_inference_visualization} further examines the ordered future sequence from $T$ to $T+100$\,ms.
The red arrows mark over-segmentation and the yellow arrows mark under-segmentation, making it clear where errors accumulate across horizons.
The sequence view shows that WeldMamba preserves the temporal evolution of the three regions more consistently, with fewer abrupt mask jumps between adjacent horizons.
In the provided sequence, the model maintains relatively consistent and stable segmentation of the wire and molten pool across multiple frames.
The remaining mistakes are still concentrated on the keyhole, whose small size and rapid motion make it the most difficult structure to predict reliably, making it an important direction for further improvement.

\section{Conclusion}
\label{sec:conclusion}

This paper presented WeldMamba, a physics-guided spatiotemporal state space network for lookahead molten pool segmentation in laser wire-feed welding.
The model connects top-view coaxial grayscale images with process parameters and aligned wire-state electrical signals, and predicts the future semantic layout of the keyhole, wire, and molten pool regions one control horizon ahead.
The main contributions are summarized as follows:

\begin{enumerate}
\item A compact physics vector from process parameters and wire-state electrical signals conditions the image encoder so local visual contrast is interpreted relative to the active welding state.
A patch-level temporal state space network with residual Mamba-2 blocks and shifted-window communication aggregates dense features across a short, non-uniform visual history, with a previous-mask marker that supplies explicit short-term semantic memory.
Operating on dense feature maps rather than per-frame vectors preserves the small keyhole and wire-tip structures, and the multi-time-scale history matches the welding process, where dense recent frames carry keyhole and wire-transfer dynamics while a longer non-uniform span captures molten pool deformation and heat accumulation.

\item A horizon-conditioned query predicts both a future latent state and a dense future feature map from historical visual tokens, process context, and sensor dynamics, allowing the decoder to infer the target future layout rather than copying the anchor-frame mask.
The future-token cross-attention decoder further aligns the predicted latent state with spatial feature tokens, while the keyhole-motion branch provides an explicit motion prior for the smallest and fastest target region, improving anticipation of keyhole displacement under beam oscillation, recoil pressure, and wire-pool interaction.

\item Experiments on the 43-sequence welding dataset show that WeldMamba reaches 74.63\% mIoU at a 500 ms lookahead, and the component ablation identifies historical frames, PatchTemporalSSM, future prediction, and keyhole motion awareness as the main contributors.
The Grad-CAM visualization further shows that the visual backbone attends to the wire-entry region, keyhole neighborhood, and molten pool interior under varying illumination and interference conditions.
\end{enumerate}

Future studies will validate the method across additional welding platforms, materials, and sensing configurations, and will connect the predicted future masks to closed-loop control experiments.
Another direction is to incorporate uncertainty estimation or richer electrical/thermal signals so that the model can express low-confidence predictions during abrupt wire-transfer and keyhole-instability events.
Deployment-oriented work will also complete the ONNX/TensorRT benchmark so that latency, memory use, and segmentation accuracy can be reported together.

\section*{Declarations}
\paragraph{Ethics approval} Not applicable.
\paragraph{Consent to participate} The authors declare their consent to participate.
\paragraph{Consent for publication} The authors declare their consent for publication.
\paragraph{Conflict of interest} The authors declare no competing interests.

\section*{CRediT Authorship Contribution Statement}
Sen Li: Writing -- review \& editing, Writing -- original draft, Visualization, Validation, Software, Methodology, Investigation, Formal analysis, Data curation.
Changhao Yin: Writing -- review \& editing, Resources, Project administration, Funding acquisition, Data curation, Conceptualization.
Chendong Shao: Investigation, Data curation.
Yaqi Wang: Resources.
Xinhua Tang: Resources, Investigation.
Haichao Cui: Supervision, Resources.

\section*{Funding}
The authors gratefully acknowledge the financial support by the State Key Research and Development Program (No. 2023YFB3407800), National Natural Science Foundation of China (No. 52575416 and No. 52305390), and Shanghai Science and Technology Commission (No. 24TS1413700, No. 24TS1414200, No. 25TS1412900, and No. 23XD1401700).

\end{document}